\pdfoutput=1
\documentclass{article}

\usepackage{microtype}
\usepackage{graphicx}
\usepackage{subfig}
\usepackage{booktabs} % for professional tables

\usepackage{hyperref}

\usepackage[accepted]{mlsys2025}

\usepackage{amsmath}
\usepackage{amssymb}
\usepackage{mathtools}
\usepackage{amsthm}
\usepackage[capitalize,noabbrev]{cleveref}

\usepackage[utf8]{inputenc} % allow utf-8 input
\usepackage[T1]{fontenc}    % use 8-bit T1 fonts
\usepackage{hyperref}       % hyperlinks
\usepackage{url}            % simple URL typesetting
\usepackage{booktabs}       % professional-quality tables
\usepackage{amsfonts}       % blackboard math symbols
\usepackage{nicefrac}       % compact symbols for 1/2, etc.
\usepackage{microtype}      % microtypography
\usepackage{xcolor}         % colors
\usepackage{wrapfig}
\usepackage{amsmath}
\usepackage{amssymb}
\usepackage{amsthm}
\usepackage{booktabs}

\usepackage{enumitem}
\usepackage{makecell}
\usepackage{diagbox}
\usepackage{bm}
\usepackage{dsfont}
\usepackage{subfig}
\usepackage{multirow}
\usepackage{varwidth}
\usepackage{pifont}
\usepackage{makecell}
\usepackage{comment}
\usepackage{color}
\usepackage[colaction]{multicol}
\usepackage{colortbl}
\usepackage{algorithm}%%%Zhao: I added this, don't delete it.
\usepackage{xspace}
\usepackage{rotating}

\usepackage{adjustbox}
\usepackage{threeparttable}
\usepackage{listings}

\theoremstyle{plain}
\newtheorem{theorem}{Theorem}[section]

\theoremstyle{definition}

\theoremstyle{remark}

\usepackage{enumitem}
\usepackage{amssymb}

\let\classAND\AND
\let\AND\relax
\let\classOR\OR
\let\OR\relax
\usepackage{algorithmic}
\let\AND\classAND

\let\OR\classOR
\AtBeginEnvironment{algorithmic}{\let\AND\algoAND}
\AtBeginEnvironment{algorithmic}{\let\OR\algoOR}

\floatname{algorithm}{Algorithm}
\renewcommand{\algorithmicrequire}{\textbf{Input:}}
\renewcommand{\algorithmicensure}{\textbf{Output:}}
\renewcommand{\algorithmicendif}{\textbf{end}}

\definecolor{gc_pink}{HTML}{FF6F79}
\definecolor{gc_blue}{HTML}{6FB0FF}
\definecolor{gc_gray}{HTML}{D9D9D9}
\definecolor{gc_dark_pink}{HTML}{99262E}
\definecolor{gc_dark_blue}{HTML}{265A99}
\definecolor{gc_dark_gray}{HTML}{999999}
\definecolor{comment_color}{HTML}{1B8F44}
\definecolor{comment_color_2}{RGB}{64,128,128}

\newcommand{\method}{MTraining}
\newcommand{\LineComment}[1]{\vspace*{0.5em}\small\textcolor{comment_color_2}{\textit{\# #1}}}

\definecolor{codegreen}{rgb}{0,0.6,0}
\definecolor{codegray}{rgb}{0.5,0.5,0.5}
\definecolor{codepurple}{rgb}{0.58,0,0.82}
\definecolor{backcolour}{rgb}{0.95,0.95,0.92}

\lstdefinestyle{mystyle}{
    backgroundcolor=\color{backcolour},   
    commentstyle=\color{codegreen},
    keywordstyle=\color{magenta},
    numberstyle=\tiny\color{codegray},
    stringstyle=\color{codepurple},
    basicstyle=\ttfamily\footnotesize,
    breakatwhitespace=false,         
    breaklines=true,                 
    captionpos=false,                    
    keepspaces=true,                 
    numbersep=5pt,                  
    showspaces=false,                
    showstringspaces=false,
    showtabs=false,                  
    tabsize=2
}

\mlsystitlerunning{MTraining: Distributed Dynamic Sparse Attention for Efficient Ultra-Long Context Training}

\begin{document}

% Add ACM Artifact Badge to the top-right of the first page
\AddToShipoutPicture*{%
  \AtTextUpperLeft{%
    \put(\LenToUnit{\textwidth-55pt},\LenToUnit{-5pt}){%
      \href{https://www.acm.org/publications/policies/artifact-review-and-badging-current}{%
        \includegraphics[width=65pt]{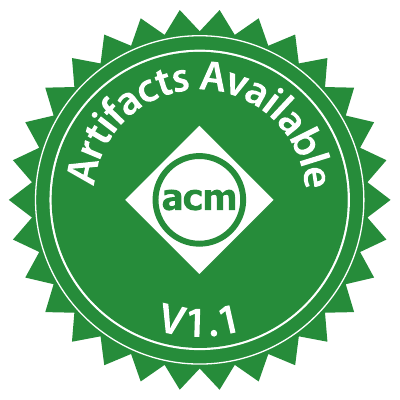}%
      }%
    }%
  }%
}

% \AddToShipoutPictureBG*{
%   \AtPageUpperLeft{
%     \hspace*{\paperwidth}
%     \raisebox{-68pt}{
%       \llap{
%         \href{https://www.acm.org/publications/policies/artifact-review-and-badging-current}{
%           \includegraphics[height=65pt]{figs/badge_artifacts-available-v1.1.pdf}}
%         \hspace{1pt}
%         % \href{https://www.acm.org/publications/policies/artifact-review-and-badging-current}{
%         %   \includegraphics[height=65pt]{figures/artifacts-functional-v1.1.pdf}}
%         % \hspace{1pt}
%         % \href{https://www.acm.org/publications/policies/artifact-review-and-badging-current}{
%         %   \includegraphics[height=65pt]{figures/results-reproduced-v1.1.pdf}}
%         \hspace{80pt}
%       }
%     }
%   }
% }

\twocolumn[
\mlsystitle{MTraining: Distributed Dynamic Sparse Attention for Efficient Ultra-Long Context Training}

\mlsyssetsymbol{equal}{*}

\begin{mlsysauthorlist}
\mlsysauthor{Wenxuan Li}{equal,msra}
\mlsysauthor{Chengruidong Zhang}{equal,msra}
\mlsysauthor{Huiqiang Jiang}{equal,msra}
\mlsysauthor{Yucheng Li}{surrey}
\mlsysauthor{Yuqing Yang}{msra}
\mlsysauthor{Lili Qiu}{msra}
\end{mlsysauthorlist}

\mlsysaffiliation{msra}{Microsoft Research}
\mlsysaffiliation{surrey}{University of Surrey}

\mlsyscorrespondingauthor{Wenxuan Li}{wl446@cantab.ac.uk}
\mlsyscorrespondingauthor{Chengruidong Zhang}{starmie.choice.specs@outlook.com}
\mlsyscorrespondingauthor{Huiqiang Jiang}{iofu728@gmail.com}

\mlsyskeywords{Machine Learning, MLSys}

\vskip 0.3in

\begin{abstract}
The adoption of long context windows has become a standard feature in Large Language Models (LLMs), as extended contexts significantly enhance their capacity for complex reasoning and broaden their applicability across diverse scenarios. Dynamic sparse attention is a promising approach for reducing the computational cost of long-context training. However, efficiently training LLMs with dynamic sparse attention on ultra-long contexts, especially in distributed settings, remains a significant challenge, largely due to worker- and step-level imbalance. This paper introduces MTraining, a novel distributed methodology leveraging dynamic sparse attention to enable efficient training for LLMs with ultra-long contexts. Specifically, MTraining integrates three key components: a distributed sparse index approximating algorithm, balanced sparse ring attention, and hierarchical sparse ring attention. These components are designed to synergistically address the computational imbalance and communication overheads inherent in dynamic sparse attention mechanisms during training LLMs with extensive context lengths. We demonstrate the efficacy of MTraining mainly by training Qwen2.5-3B and Llama-3.1-8B, successfully expanding its context window from 32K/128K to 512K tokens on a cluster of 32$\times$ A100 GPUs. Our evaluations on a comprehensive suite of downstream tasks, including RULER, PG-19, InfiniteBench, and NIAH, reveal that MTraining achieves up to a 6x higher training throughput while preserving model accuracy. The core code is available at \href{https://github.com/microsoft/MInference/tree/main/mtraining}{https://github.com/microsoft/MInference/tree/main/mtraining}.
\end{abstract}
]



\printAffiliationsAndNotice{\mlsysEqualContribution} % otherwise use the standard text.

\section{Introduction}
Long-context modeling capability is increasingly recognized as a critical capability for next-generation Large Language Models (LLMs). Many emergent applications, including long document understanding \cite{caciularu2023peek, ma2024mmlongbench}, repository-level code analysis \cite{jimenez2023swe-bench, jain2025livecodebench}, autonomous agent systems~\cite{deepresearch,manus}, and long chain-of-thought reasoning \cite{guo2025deepseek-r1,qwen3}, require LLMs to process sequences spanning hundreds of thousands to even millions of tokens.
% This extended context enables in-depth reasoning, captures long-range dependencies, and supports robust performance on complex tasks. 

\begin{figure}[!tp]
  % \vspace{-15pt}
  \centering
  \subfloat[Heavy cost.]{
    \label{sfig:lat_scaling}
    \includegraphics[height=0.5\columnwidth]{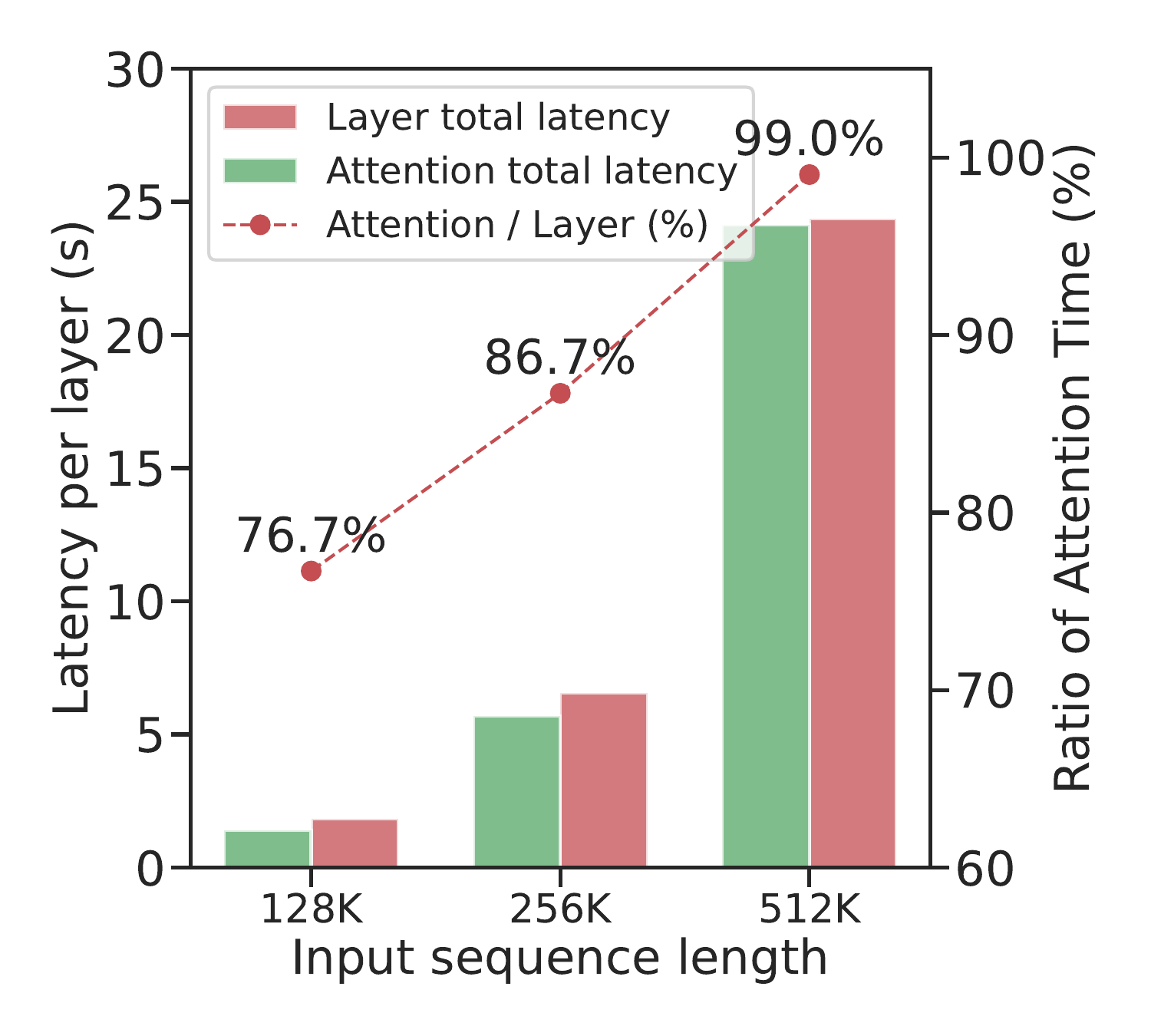}}
  \subfloat[Dynamic.]{
    \label{sfig:dynamic}
    \includegraphics[height=0.47\columnwidth]{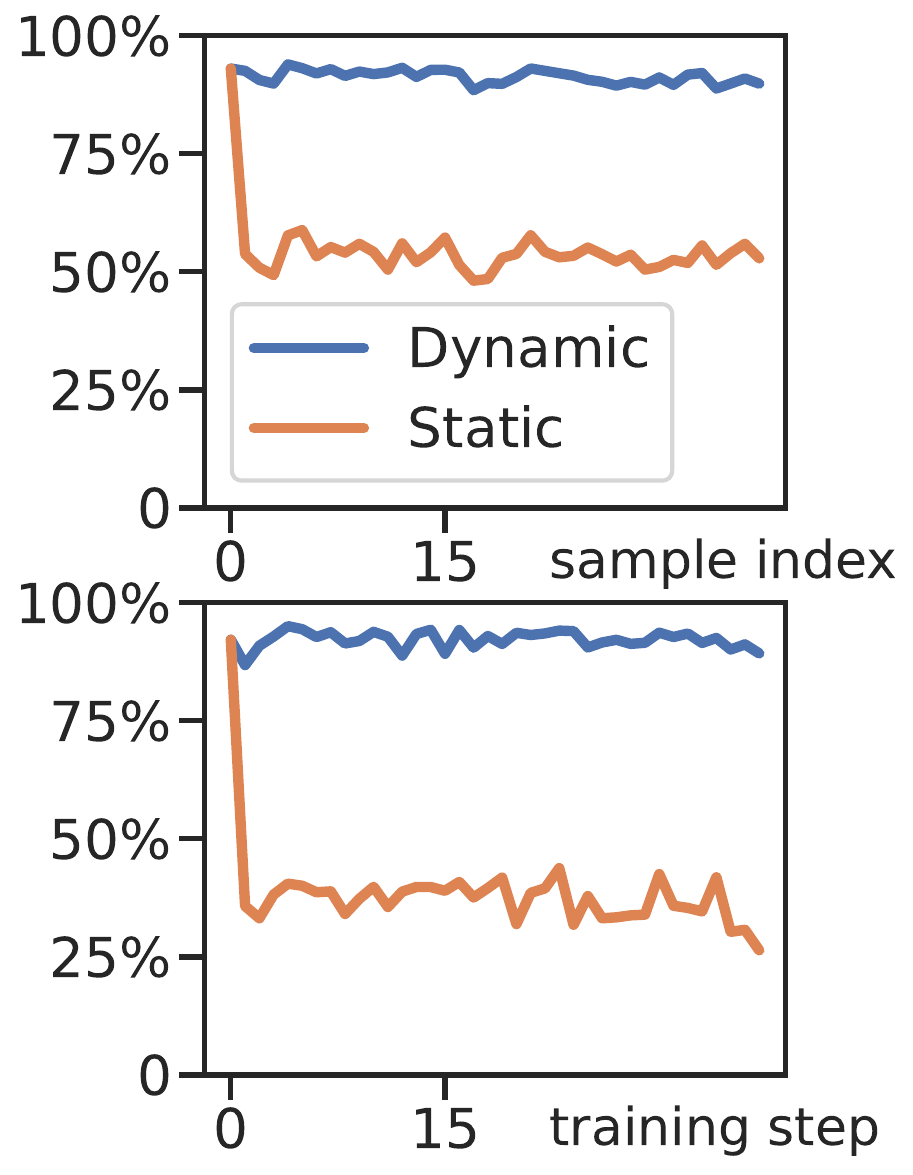}}
    
  \subfloat[Forward.]{
    \label{sfig:attention_pattern}
    \includegraphics[height=0.47\columnwidth]{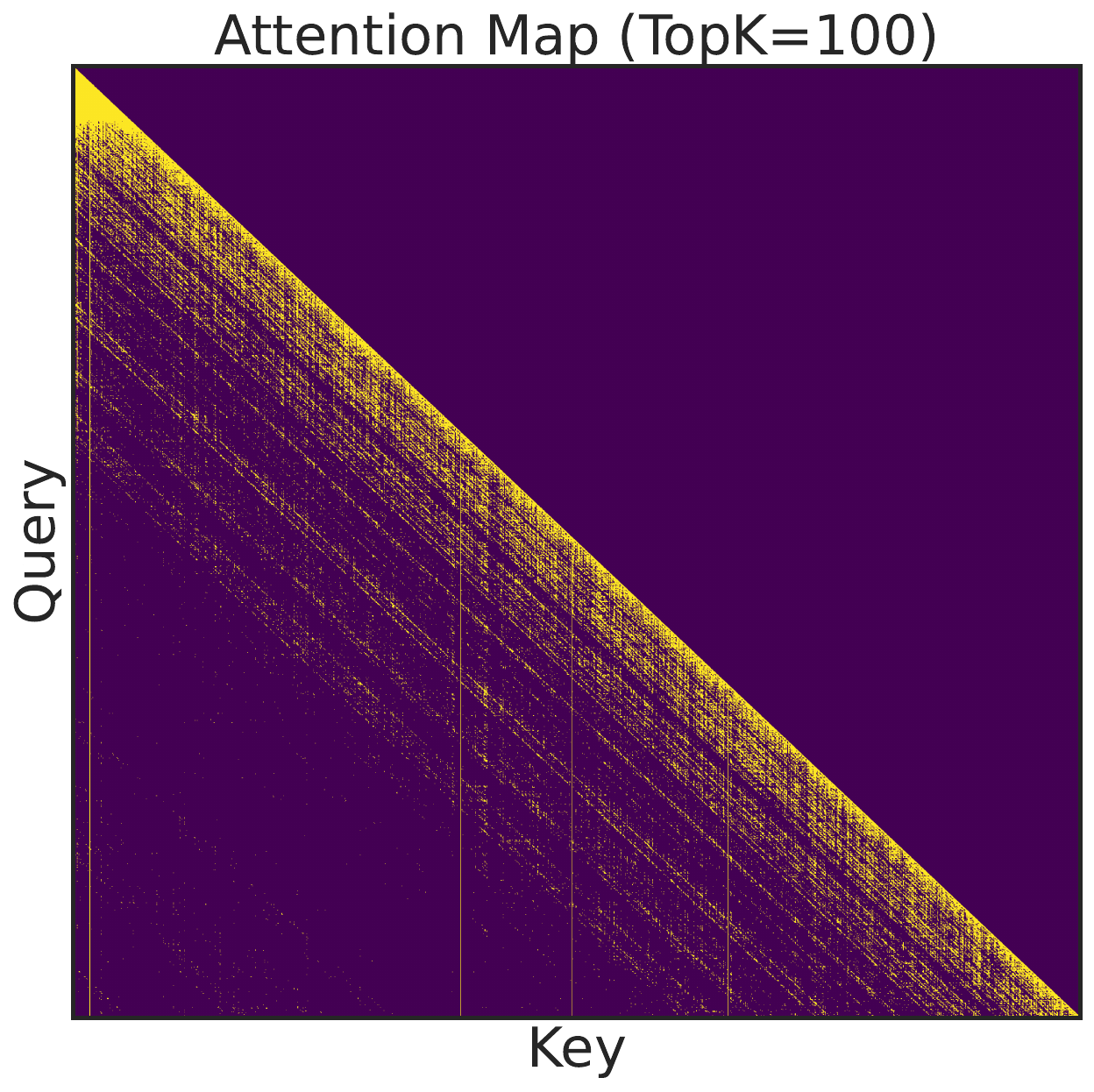}}
  \subfloat[Backward.]{
    \label{sfig:grad_pattern}
    \includegraphics[height=0.47\columnwidth]{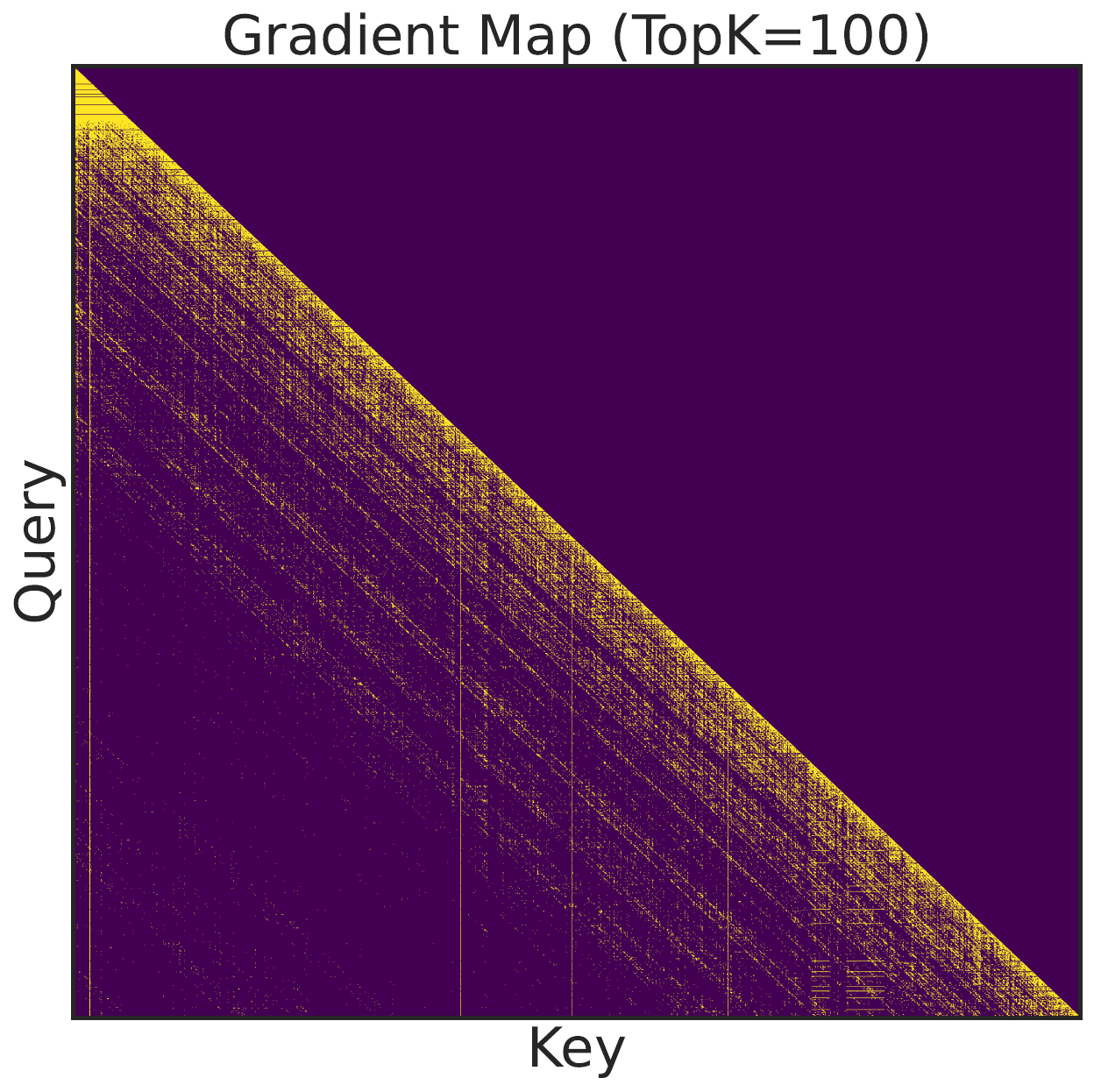}}
    % \hspace{0.2em}
    % \hspace{4em}
  \caption{ %  Structural and Clarity Improvements
  (a) Latency breakdown of the training stage: attention forward and backward dominate beyond 300K tokens.
  (b) Attention recall of top-$k$ ($k{=}1024$) entries from 128K context across different samples and training steps, illustrating the dynamic, sample-dependent nature of attention sparsity.
  (c--d) Visualization of attention weights (c) and their gradients (d) during training, revealing the Vertical-Slash (VS) pattern, where the most significant attention entries are concentrated along discrete ``slash'' diagonal bands and ``vertical'' columns (See §\ref{subsec:sparse_pattern} for more details).}
  \label{fig:motivations1}
  \vspace{-10pt}
\end{figure}

Therefore, training LLMs on long-context inputs via continued pretraining or supervised finetuning has become a vital trend to extend LLMs' context window lengths \cite{liu2024deepseek-v3, yang2025qwen1M, grattafiori2024llama3, gao2024prolong}. However, the quadratic complexity of attention computation with respect to input sequence length can result in overwhelming computational costs when the sequence length scales up. 
As shown in Fig.~\ref{sfig:lat_scaling}, when the context exceeds 300K tokens, attention's forward and backward passes account for over 90\% of \emph{per-layer computation time}. Separately, prior work~\cite{liu2024deepseek-v3,yang2025qwen1M} (e.g., DeepSeek-V3) reports that the long-context extension \emph{training phase} alone (training on 20B tokens with 128K context windows) consumes $\sim$5\% of total pretraining GPU resources, a fraction that grows further with longer target context length and larger datasets.

Prior work has shown that attention matrices exhibit strong dynamic sparsity, motivating the development of dynamic sparse attention methods~\cite{tang2024quest, jiang2024minference, lai2025flexprefill, ribar2023sparq} to reduce the cost of long-context processing. Recently, NSA~\cite{yuan2025nsa} and MoBA~\cite{lu2025moba} extended dynamic sparse attention to the pretraining phase, achieving significant efficiency gains with minimal accuracy loss. To scale long-context training across distributed clusters, Context Parallel~\cite{liu2024ring-attention} has become a vital approach to partition the activation along the sequence dimension across workers, enabling efficient memory and computation scaling. Nevertheless, integrating dynamic sparsity into Context Parallel remains challenging, due to the worker- and step-level imbalance (\S\ref{subsec:imbalanced}), which results in actual latency speedups falling far short of their theoretical potential.

The key to reducing training latency for long-context LLMs in distributed settings is to evenly distribute activated computations across workers and steps.

Building on this insight, we propose \textbf{\method{}}, a technique that enables linear scaling of dynamic sparse attention in distributed settings, significantly accelerating the training of long-context LLMs. \method{} is an algorithm–system co-design framework that integrates a training-oriented sparse attention algorithm with a sparsity-aware context parallelism strategy.
First, we empirically observe and theoretically validate that attention weights with RoPE exhibit a distinctive Vertical-Slash locality pattern (\S\ref{subsec:sparse_pattern}). Leveraging this, we introduce an online approximate sparse budget mechanism to dynamically adapt the sparsity pattern during training (\S\ref{subsec:dynamic_sparse_training_pattern}).
Second, \method{} incorporates a block-level balanced sparse ring attention mechanism based on Striped Ring Attention~\cite{brandon2023stripe}, aligning with the observed sparsity structure to achieve worker- and step-level balance (\S\ref{subsec:balanced_sparse_ring}).
Finally, \method{} employs a Hierarchical Sparse Ring Attention design to further reduce communication overhead in heterogeneous distributed networks (\S\ref{subsec:hierarchical_balanced}).

We evaluate \method{} by continued pretraining Qwen2.5-3B~\cite{yang2024qwen2-5} and Llama-3.1-8B-Instruct on ProLong~\cite{gao2024prolong} to extend their context windows from 32K/128K to 512K. On 32 NVIDIA A100-40 GB GPUs, \method{} delivers near-linear scaling of the training efficiency with dynamic sparse attention, achieving up to a 6x throughput speed-up over dense attention and 2.6x over a naïve distributed sparse attention baseline for 512K-token contexts. The resulting models are assessed on a range of long-context benchmarks with sequence lengths up to 512K tokens, including RULER~\cite{hsieh2024ruler}, Needle In A Haystack~\cite{kamradt2023needle}, InfiniteBench~\cite{zhang2024InfiniteBench}, and PG-19~\cite{rae2019pg19}. The evaluation results validate that the model trained by \method{} successfully extends its context window to 512K with matching or superior performance on these benchmarks compared to the baseline.

\section{Preliminary}
\label{sec:preliminary}
\paragraph{Ring Attention}

% \begin{wrapfigure}{r}{0.55\columnwidth}
%     \vspace{-12pt}
%     \centering
%     \includegraphics[width=\linewidth]{figs/stripe_zigzag_static.pdf}
%     \caption{Workload distribution over 4 CP workers (GPUs) in Striped and Zigzag Ring Attention. }
%     \label{fig:stripe_zigzag_static}
%     % \vspace{-12pt}
% \end{wrapfigure}

\begin{figure}[h]
    % \vspace{-8pt}
    \centering
    \includegraphics[width=0.96\linewidth]{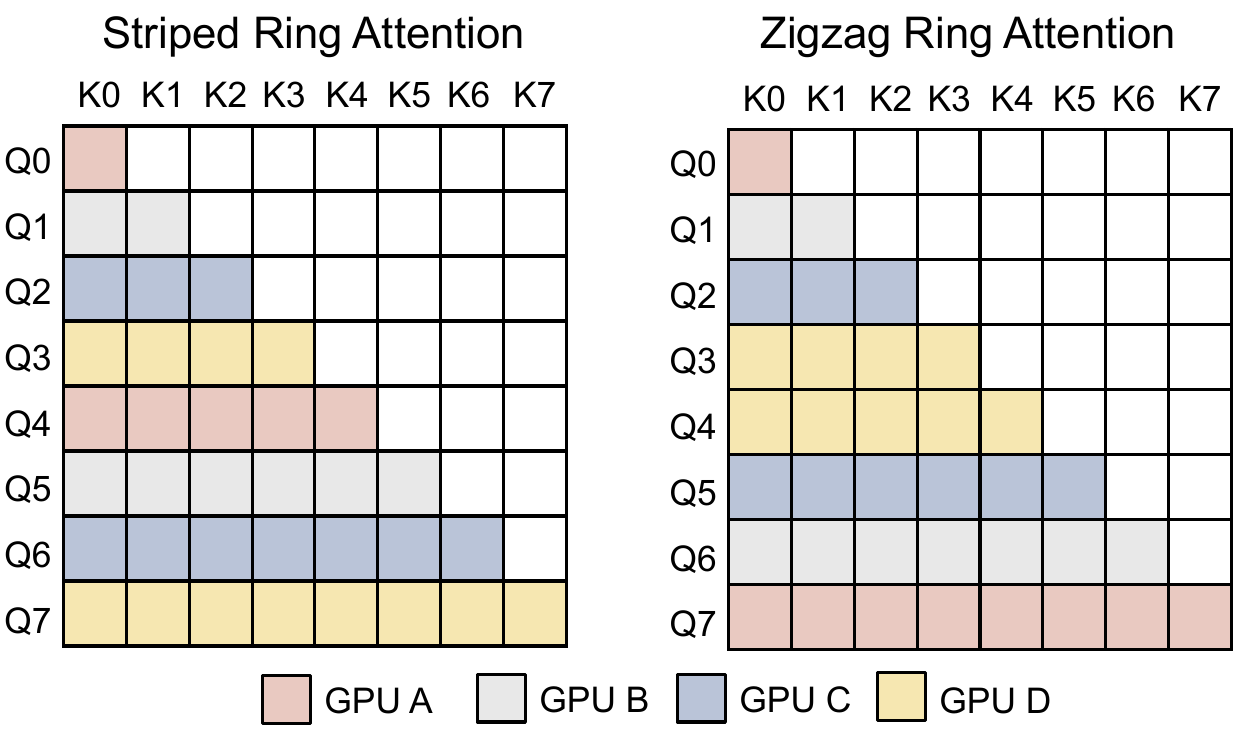}
    \caption{Workload distribution over 4 CP workers (GPUs) in Striped and Zigzag Ring Attention. }
    \label{fig:stripe_zigzag_static}
    % \vspace{-8pt}
\end{figure}

Long-context training is increasingly bottlenecked by attention latency. Ring Attention~\cite{liu2024ring-attention, brandon2023stripe} improves scalability by distributing long sequences across devices and overlapping key–value communication with blockwise attention computation~\cite{dao2022flashattention}, allowing sequence length to scale with the number of devices.

Two main variants exist: ZigZag~\cite{zhuzilin2024zigzag} and Striped~\cite{brandon2023stripe}. As shown in Fig.~\ref{fig:stripe_zigzag_static}, ZigZag folds the query dimension and mirrors blocks across workers, while Striped partitions queries cyclically by row or block. During computation, $Q$ and $O$ remain fixed per worker, while $K$ and $V$ are circulated via P2P communication—essential for Grouped Query Attention. Under causal full attention, both variants maintain balanced workload across workers.

% \vspace{-10pt}

\paragraph{Sparse Attention}
Standard attention computes dense pairwise interactions between all tokens in a sequence, resulting in a quadratic $O(N^2)$ computational and memory cost as the sequence length $N$ increases. The observation \cite{liu2022dynamic, deng2024nature-sparse} that attention maps are often highly sparse, especially at large $N$, motivates the design of \textbf{\textit{sparse attention methods}} to accelerate attention computation. In general, these methods define sparse patterns to identify the most active regions in the attention map and skip computation on negligible entries. Early approaches employed static or clustered sparsity patterns \cite{beltagy2020longformer, zaheer2020bigbird, kitaev2020reformer}, which are simple to implement but usually require pretraining models from scratch. More recent methods introduce dynamic sparse attention \cite{jiang2024minference, zhang2025spargeattn, xu2025xattention, tang2024quest}, which selects active attention entries online according to the data distribution. Some recent works \cite{lu2025moba, yuan2025nsa} have even extended dynamic sparse attention to LLM pretraining. However, few prior works have examined how the resulting dynamic sparsity affects system efficiency, particularly in distributed LLM training environments.
% \section{Validation of Vertical-Slash Pattern}
% \subsection{Thoretical Analysis}

% \subsection{Empirical Illustration}

\section{Motivation}
\label{sec:motivation}

% \todo{Figure: context windows v.s. training latency breakdown} -> Wenxuan

% TODO: Include other figures within the same figure
% \begin{figure}[h]
%     \centering
%     \includegraphics[width=0.5\linewidth]{figs/latency_scaling.pdf}
%     \caption{Scaling of Self-Attention Latency and its ratio with respect to the end-to-end layer latency of Qwen2.5-3B during training.}
%     \label{fig:lat_scaling}
% \end{figure}

% \begin{figure*}[htp]
%   \centering
%   \subfloat[Attention incurs heavy cost.]{
%     \label{sfig:lat_scaling}\includegraphics[height=0.25\columnwidth]{figs/latency_scaling.pdf}
%     }
%   \subfloat[Dynamic.]{
%     \label{sfig:dynamic}
%     \includegraphics[height=0.35\columnwidth]{figs/attn_recall.pdf}}
%   \subfloat[Attention forward.]{
%     \label{sfig:attention_pattern}
%     \includegraphics[height=0.25\columnwidth]{figs/attn_4k_800.pdf}}
%   \subfloat[Attention backward.]{
%     \label{sfig:grad_pattern}
%     \includegraphics[height=0.25\columnwidth]{figs/grad_4k_800.pdf}}
%     % \hspace{0.2em}
%     % \hspace{4em}
%   \caption{
%   (a) Latency breakdown of the training stage. 
%   (b) The attention recall of top-k(k=1024) from 128K context in different sample and training step.
%   (c-d) Visualization of attention weights (c) and their gradients (d) during training. Results are based on Qwen2.5-3B~\cite{yang2024qwen2-5} trained with a 4×8 A100 cluster.}
%   \label{fig:motivations1}
%   \vspace{-10pt}
% \end{figure*}

\subsection{Long-context Training is Dynamic Sparse}

\begin{figure}[tb]
  % \vspace{5pt}
  \centering
  % \subfloat[Imbalance in computation (FLOPs) across CP workers using XAttention~\cite{xu2025xattention}. Imbalance degree $=\text{max}/\text{mean}$.]{
  \subfloat[Imbalance in sparse attention workload (TFLOPs) across 8 workers using Tensor-Parallel and ZigZag Ring-Attention. Imbalance degree $:=\text{max}/\text{mean}$.]{
    \includegraphics[width=0.45\textwidth]{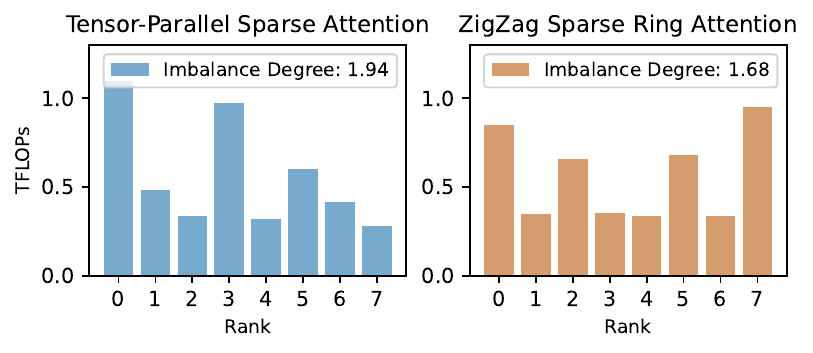}
    \label{fig:motiv_worker_imbal}
  }
  % \hfill
  \vfill
  \subfloat[Illustration of the bubble resulting from step-level imbalance, where computation and communication are not overlapped.]{
    \includegraphics[width=0.45\textwidth]{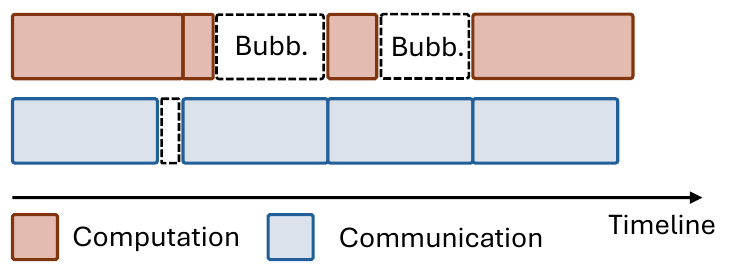}
    \label{fig:motiv_step_imbal}
  }
  \caption{Worker- and step-level load imbalance in distributed dynamic sparse attention. (a) Distribution of sparse attention FLOPs across 8 workers under Tensor Parallel (left) and ZigZag Ring Attention (right) on Qwen2.5-3B at 512k context; (b) Schematic of step-level imbalance: sparsity results in variable per-step computation time, making shorter computation steps not fully overlapped with communication.}
  \label{fig:motiv}
  \vspace{-12pt}
\end{figure}

The dynamic sparsity of attention matrices in pre-trained LLMs—especially under long-context settings—is well-documented~\cite{tang2024quest, jiang2024minference, lai2025flexprefill, xu2025xattention}. This phenomenon persists during training, often with greater variability. As shown in Fig.~\ref{sfig:dynamic}, attention sparsity fluctuates significantly across training steps and input samples.
Different model checkpoints yield distinct sparsity patterns even for the same input, reflecting temporal dynamics across training. Conversely, a single checkpoint may produce diverse sparse regions across inputs. These observations underscore the need for dynamic sparsity 
adaptation during training.

% \todo{Figure: Attention sparse distribution in training, 1) step 2) samples} -> Ruidong

\subsection{Sparse Attention Workload Exhibits Patterns}
\label{subsec:sparse_pattern}

As shown in Fig.~\ref{sfig:attention_pattern}, the attention weights display structured sparsity, consistently following a Vertical-Slash (VS) locality pattern. Geometrically, in the 2D attention matrix (query position $n$ on rows, key position $m$ on columns), a ``\emph{slash}'' corresponds to a diagonal band where $n - m = \text{const}$, capturing the locality bias induced by rotary position embeddings, while a ``\emph{vertical}'' line corresponds to a column where a particular key position $m$ receives high attention from most queries, arising from outlier tokens with unusually large key norms. Figs.~\ref{sfig:attention_pattern}--\ref{sfig:grad_pattern} visualize these two complementary structures in both the forward attention weights and their backward gradients. We further attribute the emergence of this pattern to the use of relative position embeddings, specifically RoPE~\cite{su2024rope}.
% \textbf{TODO: bridge the above and the following analysis:}
% Inspired by the definition of relative position embeddings and the visualization of RoPE~\cite{su2024rope}, we demonstrate that the attention weights exhibit a Vertical-Slash coverage pattern when it is applied with RoPE.
Let the query vector $q_n \in \mathbb{R}^{1 \times d}$ and key vector $k_m \in \mathbb{R}^{1 \times d}$ denote the token representations at positions $n, m \in \{0, \dots, N{-}1\}$ in a sequence of length $N$. We define $z_{n,m}$ as the dot product between the RoPE-transformed query and key vectors at positions $n$ and $m$, respectively.

\begin{theorem}
% The vertical-slash (VS) pattern is universally existing in RoPE-equipped LLM's self-attention weights.
The expectation of the attention weights after applying RoPE depends solely on the relative position $n-m$, i.e., $E[z_{n,m}] = \sum_{i=0}^{d-1} {\phi_{n-m}^{(i)}~ A_i} + \sum_{i=0}^{d-1} {\psi_{n-m}^{(i)}~ B_i}.$
\label{theorem:rope_slash}
\end{theorem}

\noindent Here $\phi_{n-m}^{(i)} = \cos((n-m)\theta_{i\%\frac{d}{2}})\quad$ and $\psi_{n-m}^{(i)} = (-1)^{\mathds{1}[i \ge d/2]}\sin\!\bigl((n{-}m)\,\theta_{i\% \frac{d}{2}}\bigr)$ are trigonometric basis functions determined solely by the relative position $n{-}m$ through the RoPE frequency $\theta_i = 10000^{-2i/d}$, and $A_i$, $B_i$ are position-independent constants arising from the statistical moments of query and key distributions.

Based on Theorem~\ref{theorem:rope_slash}, as proved in Appendix~\ref{appendix:proof}, we derive two key insights:
1) \textbf{Attention matrices with RoPE exhibit a Vertical-Slash coverage pattern.} 
The "slash" structure arises from the dependence of expected attention weights on the relative position $n - m$, while the "vertical" component results from outliers in the key distributions, as described in Eq.~\ref{eq:fluctuation};
2) \textbf{Attention matrices with RoPE tend to form banded sparse activation patterns.}    
Since $\phi_{n-m}^{(i)}$ and $\psi_{n-m}^{(i)}$ are continuous in the relative position $n - m$, and the coefficients $A_i$ and $B_i$ in $\mathbb{E}[z_{n,m}]$ are position-independent, activations tend to cluster locally around specific relative positions.

% \todo{Figure: Attention sparse pattern in training} -> Wenxuan

% \begin{figure}[h]
%     \centering
%     \includegraphics[width=0.5\linewidth]{figs/attn_grad_4k_800.pdf}
%     \caption{Visualization of Top-100 values of each row in the attention map and its corresponding gradients from a random sample during training.}
%     \label{fig:motiv_attn_latency}
% \end{figure}

In the backward pass, the gradient $\frac{\partial \mathcal{L}}{\partial \bm{S}}$ exhibits sparse patterns that closely mirror those in the forward pass, as shown in Fig.~\ref{sfig:grad_pattern}. Based on the attention computation equations, we can derive the gradients of the attention weights ($\bm{S} = \bm{Q}\bm{K}^\top / \sqrt{d_h}$, $ \bm{A} = \text{softmax}(\bm{S})$) as well as those of $\bm{Q}$, $\bm{K}$, and $\bm{V}$, as shown in Eq.~\ref{eq:softmax_gradient} and Eq.~\ref{eq:attention_backward}.
\begin{equation}
\begin{aligned}
    % dV &= A dO\\ 
    \frac{\partial \mathcal{L}}{\partial \bm{S}} &= \bm{A} \odot \left(\frac{\partial \mathcal{L}}{\partial \bm{A}} - \sum_j \frac{\partial \mathcal{L}}{\partial \bm{A}_{ij}} \bm{A}_{ij}\right) \\
\end{aligned}
\label{eq:softmax_gradient}
\end{equation}

% wwwwwwwhere $S = QK^\top / \sqrt{d_k}$, and $ A = \text{softmax}(S)$.

% \begin{equation}
% \begin{aligned}
%     % dV &= A dO\\ 
%     \frac{\partial \mathcal{L}}{\partial V} = A^\top \cdot \frac{\partial \mathcal{L}}{\partial \text{O}},\quad
%     \frac{\partial \mathcal{L}}{\partial Q} = \frac{1}{\sqrt{d}} \cdot \frac{\partial \mathcal{L}}{\partial S} \cdot K, \quad
%     \frac{\partial \mathcal{L}}{\partial K} = \frac{1}{\sqrt{d}} \cdot \left(\frac{\partial \mathcal{L}}{\partial S}\right)^\top \cdot Q
% \end{aligned}
% \label{eq:attention_backward}
% \end{equation}

By substituting $\frac{\partial \mathcal{L}}{\partial \bm{S}}$ into the gradient expression of attention (Eq.~\ref{eq:attention_backward}), we observe that all matrix operations (i.e., GEMMs) in the backward pass depend on the attention weights $A$. Consequently, the dynamic sparsity in the backward pass can be viewed as a superposition of the forward-phase sparsity.
% REVISION: Structural and Clarity Improvements
\begin{figure*}[t!]
    % \vspace{-5pt}
    \centering
    \includegraphics[width=0.95\linewidth]{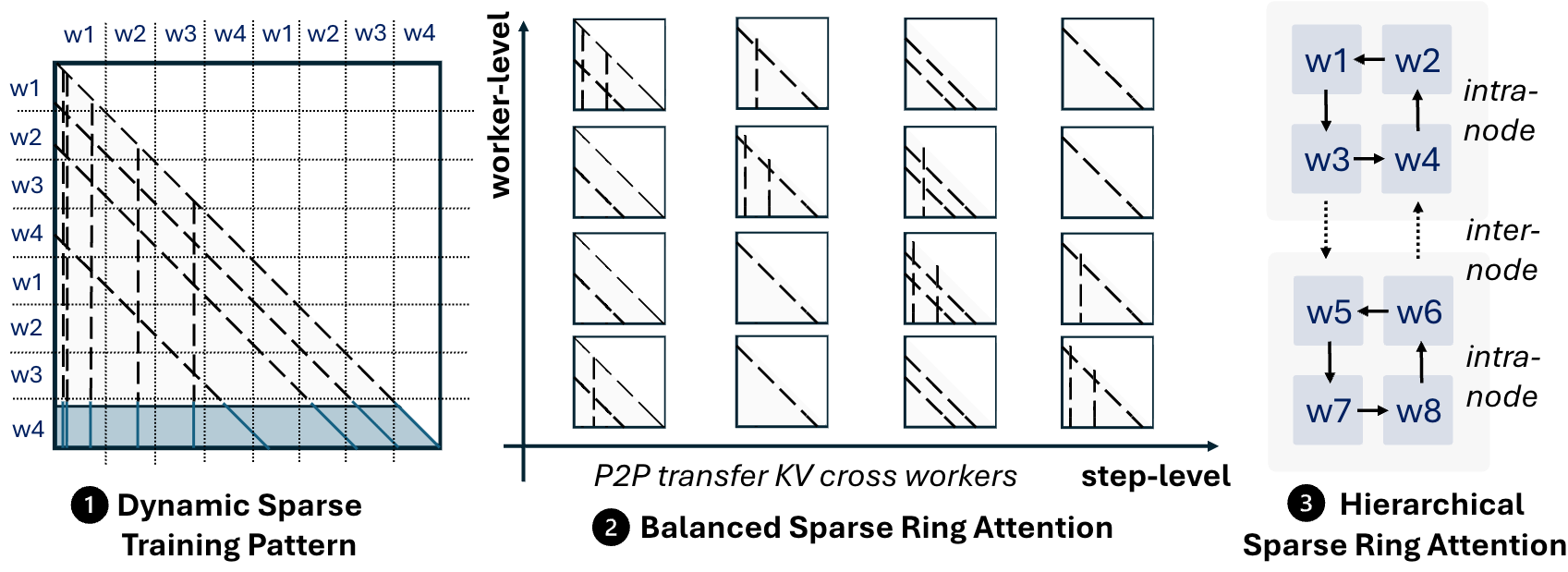}
    \caption{Overview of \method in distributed scenarios. At each training iteration the pipeline proceeds with three components: \textcircled{\small 1} \emph{Distributed Sparse Index Approximating} firstly gathers all keys and obtains statistics over an approximate attention area to estimate active attention regions. With the derived sparse mask, Ring Attention launches with \textcircled{\small 2} \emph{Balanced Sparse Ring Attention} distributing the resulting workloads via a block-level striped layout aligned with the Vertical-Slash pattern; and \textcircled{\small 3} \emph{Hierarchical Sparse Ring Attention} overlapping inter-node KV communication with intra-node ring computation.}
    \vspace{-10pt}
    \label{fig:framework}
\end{figure*}

\subsection{Distributed Dynamic Sparse Attention is Imbalanced}
\label{subsec:imbalanced}

Distributed dynamic sparse attention introduces new challenges absent in single-node dense settings—most notably, worker- and step-level imbalance.
As shown in Fig.~\ref{fig:motiv_worker_imbal}, we measure the balance of attention compute across workers under Tensor Parallel (Left) and ZigZag Ring Attention (Right, ~\cite{zhuzilin2024zigzag}) on an 8-GPU node. In both cases, dynamic sparsity leads to uneven FLOPs across workers, causing faster workers to idle at synchronization barriers. These results highlight that neither traditional TP nor naively applied CP can maintain balanced workloads under dynamic sparsity, with imbalance degree reaching 1.94 with TP.

% underscoring the need for a balanced distributed sparse attention mechanism.

% Distributed dynamic sparse attention introduces new challenges absent in single-node settings—most notably, worker- and step-level imbalance.
% As shown in Fig.~\ref{fig:motiv_worker_imbal}, the dynamic sparsity leads to uneven FLOPs across workers, causing worker-level imbalance where faster workers idle due to synchronization barriers.
% We define the imbalance degree as the ratio of the maximum to average per-worker FLOPs in a step. 
% For example, with XAttention~\cite{xu2025xattention} at 95\% sparsity and 32-way context parallelism, the imbalance degree reaches 3.17—reducing realized speedup to one-third of the theoretical maximum.

% \begin{wrapfigure}{r}{0.45\columnwidth}
% \vspace{-10pt}
%     \centering
%     \includegraphics[width=\linewidth]{figs/motiv_worker_imbalance.pdf}
%     \caption{Imbalance in computation (FLOPs) across CP workers using XAttention~\cite{xu2025xattention}. Imbalance degree $=\text{max}/\text{mean}$.}
%     \vspace{-30pt}
%     \label{fig:motiv_worker_imbal}
% \end{wrapfigure}

% REVISION: Structural and Clarity Improvements
On the other hand, step-level imbalance refers to fluctuations in a single worker's computational load across the successive Ring Attention steps \emph{within a single training iteration}: as KV chunks circulate around the ring, each step encounters a different region of the attention matrix with a distinct sparsity ratio, leading to variable per-step computation. As shown in Fig.~\ref{fig:motiv_step_imbal}, this variation can cause uneven workloads over time. When computation is reduced due to high sparsity, it may fall below communication latency, making it harder to overlap compute and communication, leading to performance-degrading bubbles.

% \begin{wrapfigure}{r}{0.45\columnwidth}
%     \vspace{-15pt}
%     \centering
%     \includegraphics[width=\linewidth]{figs/step-level_imbalance.pdf}
%     \caption{Illustration of the bubble resulting from step-level imbalance, where computation and communication are not overlapped.}
%     \vspace{-15pt}
%     \label{fig:motiv_step_imbal}
% \end{wrapfigure}

% \begin{figure*}[htb]
%   \vspace{-10pt}
%   \centering
%   \subfloat[Imbalance in computation (FLOPs) across CP workers using XAttention~\cite{xu2025xattention}. Imbalance degree $=\text{max}/\text{mean}$.]{
%     \label{fig:motiv_worker_imbal}
%     \includegraphics[width=0.4\linewidth]{figs/motiv_worker_imbalance.pdf}}

%   % \hspace{0.1em}
    
%   \subfloat[Illustration of the bubble resulting from step-level imbalance, where computation and communication are not overlapped.]{
%     \label{fig:motiv_step_imbal}
%     \includegraphics[width=0.4\linewidth]{figs/step-level_imbalance.pdf}}
%     % 
%     % \hspace{4em}
%   \caption{Step-level Computation Schedule of Striped Ring Attention (a) and Hierarchical Striped Ring Attention (b) with 4 CP workers.}
%   \label{fig:motiv_step_imbal}
%   \vspace{-10pt}
% \end{figure*}

% \todo{Figure: Imbalanced issue} -> Wenxuan

% \section{VS-Pattern-Driven Sparse Training Algorithm}

% \section{Spasrsity-aware Hierarchical Context Parallelism}

\section{MTraining}

% \todo{Figure: MTraining Framework}

Building on the analysis in \S\ref{sec:motivation}, we propose \method{} to accelerate distributed training of ultra-long-context LLMs. \method{} comprises three components: 
1) \emph{Distributed Sparse Index Approximating}, tailored for distributed training with dynamic sparse index building; 
2) \emph{Balanced Sparse Ring Attention}, which applies a stripe-based layout to address worker- and step-level imbalance; 
3) \emph{Hierarchical Sparse Ring Attention}, which leverages hierarchical ring topology to overlap the communication on heterogeneous intra-/inter-node media in cross-node scenarios.

Figure~\ref{fig:framework} illustrates these three core components in the end-to-end workflow of \method{} for sparse attention computation: 
At the beginning of each attention layer, the \textit{Distributed Sparse Index Approximating} module (\S\ref{subsec:dynamic_sparse_training_pattern}) estimates the active attention regions by gathering keys vectors to compute an approximate attention area for building a shared sparse index. 
This index is then consumed by the \textit{Balanced Sparse Ring Attention} module (\S\ref{subsec:balanced_sparse_ring}) and \textit{Hierarchical Sparse Ring Attention} module (\S\ref{subsec:hierarchical_balanced}). \textit{Balanced Sparse Ring Attention} adopts the block-level striped layout for Ring Attention to balance the workload across works considering the sparse pattern,
while the \textit{Hierarchical Sparse Ring Attention} module (\S\ref{subsec:hierarchical_balanced}) overlaps inter-node KV transfers with intra-node ring computation by dividing the Ring Attention into hierarchical two layers in distributed training.

% \begin{figure}[t]
%     \vspace{-10pt}
%     \centering
%     \includegraphics[width=0.95\linewidth]{figs/mtraining_framework.pdf}
%     \caption{Overview of \method{} in distributed scenarios.}
%     % \vspace{-10pt}
%     \label{fig:framework}
% \end{figure}
% % \todo{Figure: MTraining Framework}

% \subsection{Dynamic Sparse Training Pattern}
% \label{subsec:dynamic_sparse_training_pattern}

% Distributed Sparse Index Approximating

\vspace{-5pt}
\subsection{Distributed Sparse Index Approximating}
\label{subsec:dynamic_sparse_training_pattern}

The Sparse Attention operator requires the block and/or column index as input. In the \emph{Context Parallelism} scenario, a new algorithm is needed to estimate the index on QKV data distributed among different workers. Building on MInference~\cite{jiang2024minference} and FlexPrefill~\cite{lai2025flexprefill}, we propose a novel distributed dynamic sparse index approximating algorithm guided by the Vertical-Slash structure for \textbf{distributed training}. As detailed in Algorithm~\ref{alg:dynamic_training}, our approach incorporates three key components:

% Motivated by both empirical observations and theoretical validation of the Vertical-Slash pattern during training (see \S\ref{subsec:sparse_pattern} and Appendix~\ref{appendix:proof}), we extend this dynamic sparse attention—originally designed for inference~\cite{jiang2024minference, lai2025flexprefill}—to the \textbf{distributed training} scenario. Building on MInference and FlexPrefill, we propose a novel distributed dynamic sparse pattern guided by Vertical-Slash structure for training. As detailed in Algorithm~\ref{alg:dynamic_training}, our approach incorporates three key components:

\textit{(i) Distributed Online Budget Approximation.}  To accommodate the dynamic variation in sparsity patterns across training steps and input samples without incurring the overhead of offline profiling, we introduce a distributed online budget approximation method under \emph{Context Parallelism}. In Ring Attention, each device is responsible for a segment of the query, key, and value along the sequence dimension. The last `last\_q` query entries will be broadcast to all devices for computing partial attention weights within an observation window, enabling each rank to compute local attention statistics—such as cumulative sums along the vertical and slash directions. These statistics are then gathered to estimate the minimal number of active vertical and slash lines required to recall a target proportion of total attention mass. The resulting sparse indices are subsequently broadcast to all ranks for later sparse attention computation in each device.

\textit{(ii) Kernel-Aware Approximation Granularity.} Since the vertical and slash components correspond to different granularities in GPU execution, we align the approximation resolution with the underlying kernel structure. Vertical lines are approximated at the token level to capture fine-grained activation locality, while slash lines are aggregated over $64\times64$ token blocks to match the tiling used in block-wise matmul kernels. This alignment minimizes index fragmentation and ensures consistency between sparsity estimation and actual GPU execution.

\textit{(iii) Ring Attention with Sparse Index}.  Once the distributed sparse index is synchronized across devices, each worker determines the active regions within the local attention submatrix according to the indices of active vertical and slash lines. Within each Ring Attention step, the worker selectively computes only the nonzero entries using customized attention kernels, with KV chunks being transferred across ranks concurrently to reach computation-communication overlap. Finally, each device completes attention computation over its assigned sequence segment with sparsity.

\textbf{Design-space considerations.} The granularity of sparse index approximation presents a trade-off between accuracy and overhead. Finer-grained indexing (e.g., per-token slash selection) captures more precise activation patterns but incurs higher profiling and synchronization costs, while coarser indexing (e.g., large block-level selection as in MoBA) reduces overhead at the expense of including more redundant computation. Our choice of token-level vertical indexing and $64\times64$ block-level slash indexing balances these factors by matching the GPU kernel tiling granularity. Alternative heuristics, such as fixed top-k budgets or offline-profiled static patterns, could eliminate the online estimation cost but sacrifice adaptivity to varying input distributions and training dynamics. The top-p mass threshold used in \method{} offers a middle ground: it automatically adjusts the sparsity budget per head and per layer based on the actual attention distribution, requiring no manual tuning while keeping the profiling overhead to under 6\% of total attention latency.

\subsection{Balanced Sparse Ring Attention}
\label{subsec:balanced_sparse_ring}

As discussed in \S\ref{sec:preliminary} and \S\ref{subsec:imbalanced}, both ZigZag and Striped implementations of Ring Attention achieve balanced computation under full attention with a causal mask. However, in dynamic sparse attention settings, their distinct activation patterns lead to worker- and step-level imbalance. As shown in Fig.~\ref{sfig:stripe_comp} and Fig.~\ref{fig:zigzag_comp}, ZigZag distributes computation along incomplete diagonal and anti-diagonal lines, while Striped distributes along complete diagonal lines. These differing spatiotemporal patterns result in differences in load balancing under dynamic, data-dependent sparsity.

\textbf{Choice of Striped over Zigzag Ring Attention.} The choice between Striped and Zigzag Ring Attention is guided by their alignment with the Vertical-Slash sparsity pattern validated in \S\ref{subsec:sparse_pattern}. As shown in Fig.~\ref{sfig:stripe_comp} and Fig.~\ref{fig:zigzag_comp}, at each Ring Attention step, Striped Ring Attention assigns workers to compute attention blocks along complete slash lines parallel to the main diagonal, whereas Zigzag Ring Attention distributes computation along mixed diagonal and anti-diagonal lines. Since the dominant sparse activations concentrate along slash lines (Theorem~\ref{theorem:rope_slash}), Striped Ring Attention naturally aligns each worker's assigned blocks with the regions of highest attention mass, yielding more uniform per-worker computation. Furthermore, Striped Ring Attention offers finer granularity control: the stripe width can be tuned down to a single block (64 tokens in our case), so that the shift in computation blocks between consecutive steps remains small, reducing step-level workload fluctuation. In contrast, Zigzag Ring Attention has a fixed granularity of $N/(2W)$ (where $N$ is the sequence length and $W$ is the number of workers), which grows with input length and introduces larger inter-step workload variations.

\restylefloat{algorithm}
% \begin{minipage}
% \begin{algorithm}[!tp]
\begin{algorithm}[!t]
    % \vspace{-10pt}
  \captionsetup[algorithm]{singlelinecheck=off} % optional
  \caption{Distributed Sparse Attention Forward}
  \label{alg:dynamic_training}
  \begin{algorithmic}
  % \vspace{8pt}
  \STATE {\bfseries Input:} $\boldsymbol{Q}^j,\boldsymbol{K}^j,\boldsymbol{V}^j \in \mathbb{R}^{T \times d_h}$ on each worker $j$, Top-P budget $p \in (0,1)$

\LineComment{Parallelized across workers}
\FOR{$j \gets 1$ to $N_{workers}$}
\STATE $ $
\LineComment{Broadcast last\_q to each worker}
    \IF{$j = N_{workers}$}
        \STATE $Q_{last} \gets Q^j_{[-last\_q:]}$
        \STATE $\mathrm{broadcast}(Q_{last}, \mathrm{to}=[1, 2, ..., N_{workers}])$
    \ENDIF
    \STATE $Q_{last} \gets \mathrm{recv}(\mathrm{from}=N_{workers})$

    \LineComment{Compute local attention statistics using last\_q}
    \STATE $\boldsymbol{\hat{A}}^j \gets \bm{M}_{\text{casual}} \odot \left( \boldsymbol{Q}_{last} \bm{K}^{j\top} \right) / \sqrt{d_h}$

    \LineComment{Gather attention statistics from each worker}
    \STATE $\mathrm{send}(\boldsymbol{\hat{A}}^j, \mathrm{to}=0)$
    \IF{$j = 1$}
        \STATE $\boldsymbol{\hat{A}} \gets \mathrm{gather}(\mathrm{from}=[1, 2, ..., N_{workers}])$
    \ENDIF

    \LineComment{Online approximation of budgets and Top-K indices}
    \IF{$j = 1$}
        \STATE $\boldsymbol{\hat{A}} \gets \mathrm{softmax}(\boldsymbol{\hat{A}})$
        \STATE $k_v \gets \mathrm{score\_cover}\left(\mathrm{sorted}\left(\mathrm{sum}_v(\boldsymbol{\hat{A}})\right), p\right)$
        \STATE $\boldsymbol{i}_v \gets \mathrm{argtopk}\left(\mathrm{sum}_v(\boldsymbol{\hat{A}}), k_v\right)$
        \STATE $k_s \gets \mathrm{score\_cover}\left(\mathrm{sorted}\left(\mathrm{sum}_s(\boldsymbol{\hat{A}})\right), p\right)$
        \STATE $\boldsymbol{i}_s \gets \mathrm{argtopk}\left(\mathrm{sum}_s(\boldsymbol{\hat{A}}), k_s\right)$
    \ENDIF

    \LineComment{Broadcast Top-K indices to each worker}
    \IF{$j = 1$}
        \STATE $\mathrm{broadcast}([i_v, i_s], \mathrm{to}=[1, 2, ..., N_{workers}])$
    \ELSE
        \STATE $[i_v, i_s] \gets \mathrm{recv}(\mathrm{from}=0)$
    \ENDIF
    
    \LineComment{Convert Top-K indices to local sparse attention index}
    \STATE $\boldsymbol{i}^j_{blk}, \boldsymbol{i}^j_{col} \gets \mathrm{convert\_index}(\boldsymbol{i}_v, \boldsymbol{i}_s, j)$

    \LineComment{Launch Sparse Ring Attention with index}
    \STATE $\boldsymbol{O}^j \gets \mathrm{sparse\_ring\_attn}(\boldsymbol{Q}^j, \boldsymbol{K}^j, \boldsymbol{V}^j, \boldsymbol{i}^j_{blk}, \boldsymbol{i}^j_{col})$
    \STATE $\mathrm{return}\,\,\,\boldsymbol{O}^j$

\ENDFOR

    % \vspace{4.5pt}
  \end{algorithmic}
  % \vspace{-9pt}//
\end{algorithm}

To address this, we propose Balanced Sparse Ring Attention, a system–algorithm co-design approach comprising the following key components:

% \begin{figure*}[htb]
%   \vspace{-10pt}
%   \centering
%   \subfloat[Balanced Stripped Sparse Ring Attention.]{
%     \label{sfig:stripe_comp}
%     \includegraphics[width=0.5\linewidth]{figs/Stripe_Comp.pdf}}
%   \subfloat[Hierarchical Balanced Sparse Ring Attention.]{
%     \label{sfig:hierarchical_comp}
%     \includegraphics[width=0.5\linewidth]{figs/DR_Stripe_Comp.pdf}}
%     % \hspace{0.2em}
%     % \hspace{4em}
%   \caption{Step-level Computation Schedule of Striped Ring Attention (a) and Hierarchical Striped Ring Attention (b) with 4 CP workers.}
%   \label{fig:comp_schedule}
%   \vspace{-10pt}
% \end{figure*}

\textit{(i) Striped Sparse Ring Attention.}
As shown in \S\ref{subsec:sparse_pattern} and \S\ref{subsec:dynamic_sparse_training_pattern}, RoPE-based attention during training predominantly exhibits a Vertical-Slash sparsity pattern, where slash components dominate the computation due to (1) slash lines originate from statistical properties and appear more frequently than vertical lines originating from outliers, and (2) Flash-Attention needs to cover slash lines with blocks on GPUs, requiring redundant computation. To balance workload across workers, we align them along the diagonal direction and propose a striped dynamic sparse ring attention scheme. As shown in Fig.~\ref{sfig:stripe_comp}, this design evenly distributes slash lines across workers, allowing each to process contiguous slash regions at each step.

\textit{(ii) Block-level Striped Sparse Ring Attention.}
Due to the block-level computation of slash operations and their spatial locality, we introduce block-level striped sparse ring attention. We adopt a 64-token stripe granularity to preserve coherence, avoid fragmentation from token-level striping, and maintain kernel sparsity and efficiency. This alignment also reduces index overhead and improves runtime performance.

\textit{(iii) Step-level Balanced Ring Attention.}
Our block-level striped design also mitigates step-level imbalance. In ultra-long-context settings, workers process fine-grained stripes at each step—for example, with 128 workers and a 512K sequence, each worker handles 64 block stripes sequentially. This repeated, fine-grained partitioning stabilizes computation across steps, ensuring more consistent workload distribution.

\vspace{-10pt}

\subsection{Hierarchical Balanced Sparse Ring Attention}
\label{subsec:hierarchical_balanced}

\begin{figure*}[t]
  % \vspace{-10pt}
  \centering
  \subfloat[Balanced Stripped Sparse Ring Attention.]{
    \label{sfig:stripe_comp}
    \includegraphics[width=0.5\linewidth]{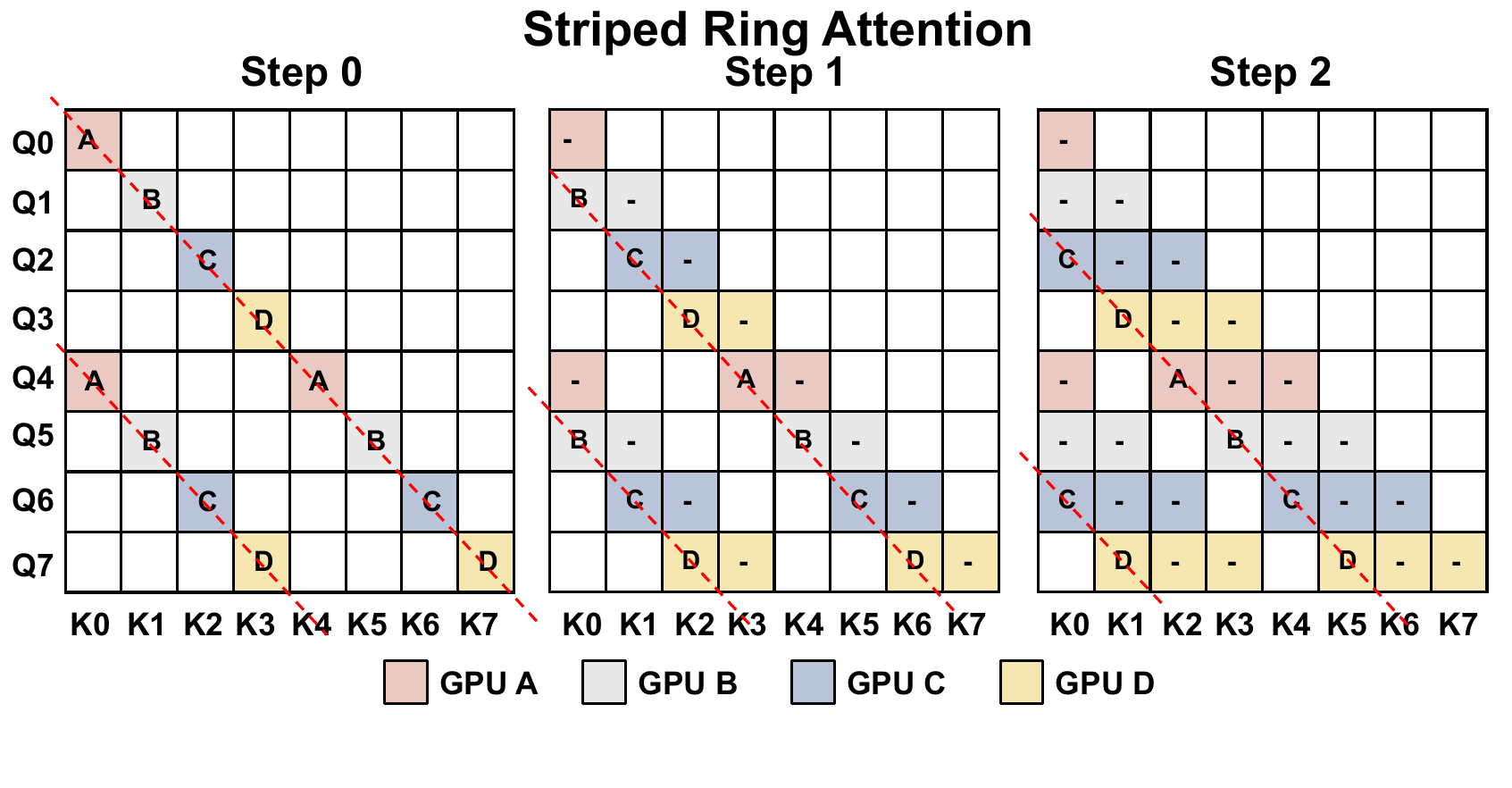}}
  \subfloat[Hierarchical Balanced Sparse Ring Attention.]{
    \label{sfig:hierarchical_comp}
    \includegraphics[width=0.5\linewidth]{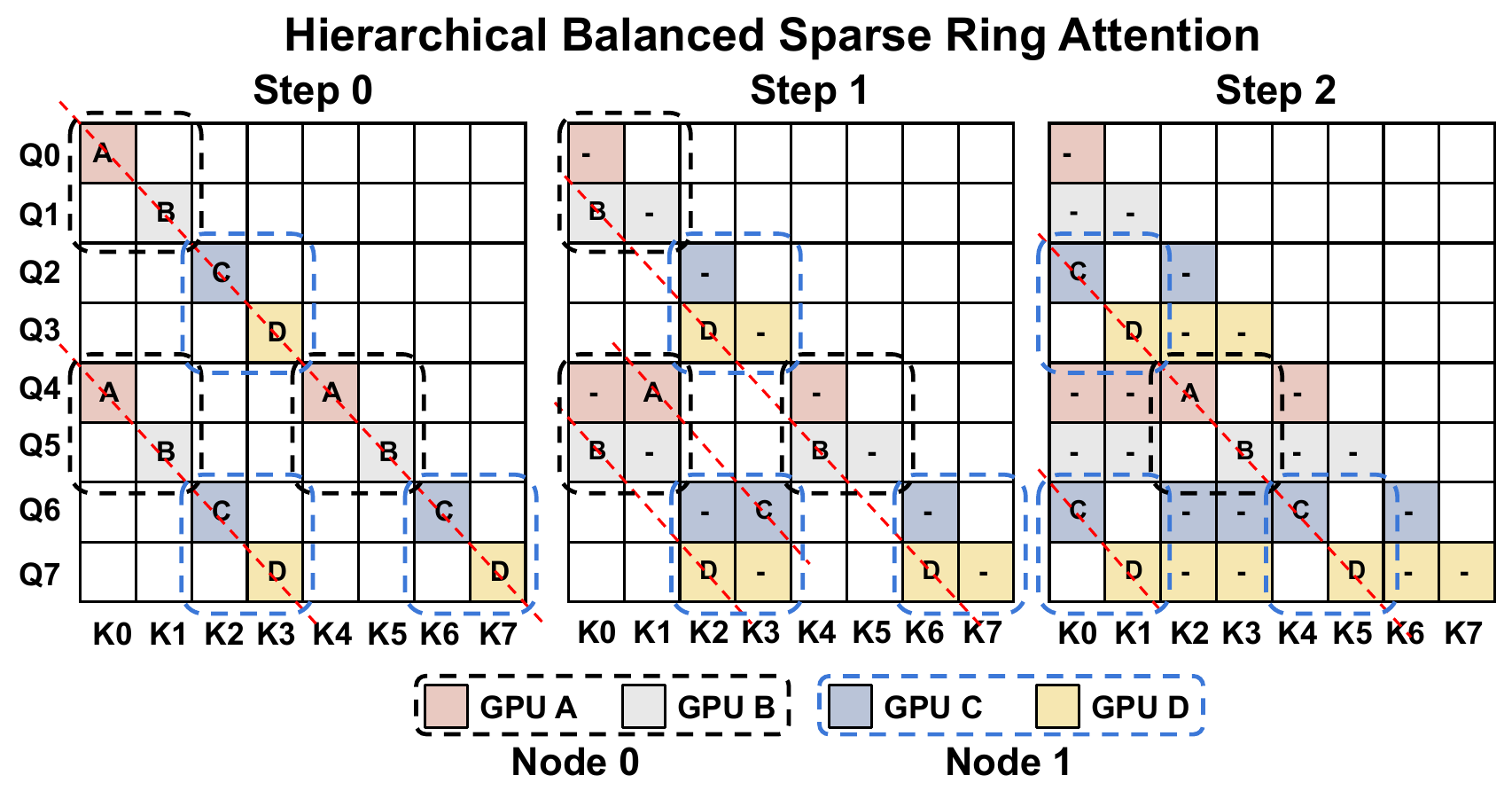}}
    % \hspace{0.2em}
    % \hspace{4em}
  \caption{Step-level computation schedule of the two Ring Attention variants with 4 CP workers (labeled A--D): (a) \textbf{Balanced Striped Sparse Ring Attention}: at each Ring Attention step (column), workers compute attention blocks along slash lines parallel to the main diagonal, aligning with the VS pattern to achieve balanced workloads; (b) \textbf{Hierarchical Balanced Sparse Ring Attention}: the inner ring circulates KV within a node while the outer ring transfers KV across nodes, where workers grouped by the same dash box are co-located on the same compute node.}
  \label{fig:comp_schedule}
  % \vspace{-10pt}
\end{figure*}

Ring Attention typically overlaps computation and communication by concurrently executing matmul and communication kernels~\cite{liu2024ring-attention}. However, with dynamic sparsity, reduced per-worker computation amplifies communication overhead, making it critical to mitigate communication cost for efficient distributed training under sparse regimes. Especially in distributed training with heterogeneous communication links, inter-node communication often becomes the bottleneck in Ring Attention. For example, inter-node bandwidth (e.g., 25 GB/s InfiniBand HDR) is typically 3–12× slower than intra-node links such as NVLink 3.0 (300 GB/s) or PCIe 5.0.
% \textbf{TODO: Here provide a real numerical example}
Recent works~\cite{liu2024deepseek-v3, gu2024loongtrain} have explored hierarchical communication topologies to reduce latency under such bandwidth asymmetry. Inspired by~\cite{gu2024loongtrain}, we propose Hierarchical Balanced Sparse Ring Attention to mitigate inter-node communication overhead in sparse ring attention.
 
In the computation schedule shown in Fig.~\ref{sfig:hierarchical_comp}, the dashed boxes group workers that reside on the same physical compute node. Within each outer-ring step, these co-located workers first complete a full round of inner-ring KV exchange and computation among themselves, while the slower inter-node transfer proceeds concurrently in the background. The temporal ordering differs from standard (flat) ring attention: instead of a single global ring that advances one worker at a time, the hierarchical schedule interleaves $N_{\text{card}}$ fast intra-node steps for every one slow inter-node step, effectively amortizing the cross-node latency over multiple local computation rounds.

Specifically, as shown in Fig.~\ref{sfig:hierarchical_comp}, our approach incorporates the following design:

\textit{(i) Inner- and Outer-Ring Hierarchical Ring Attention.}
We decompose the global ring communication into two hierarchical levels: an inner ring and an outer ring. In the inner ring, key–value (KV) blocks are circulated among the $N_{\text{card}}$ GPUs within each compute node. The outer ring handles communication across $N_{\text{node}}$ nodes by exchanging aggregated KV buffers.
At each outer-ring step, the schedule proceeds as follows: 
1) \textbf{Post Outer P2P.} A non-blocking P2P communication operation is initiated, transmitting the current KV chunk of the local node to the next node and posting a matching receive. 
2) \textbf{Inner-Ring Attention.} While the inter-node transfer is in progress, the GPUs enter a loop of length $N_{\text{card}}$, performing sparse ring attention computations over the local KV slices within the node.
3) \textbf{Asynchronous Waiting.} At the end of each outer step, wait for both computation and communication to end before moving to the next outer-ring iteration.

\textit{(ii) Hierarchical Balanced Sparse Ring Attention.}
Applying hierarchical ring attention in the sparse setting alters the propagation order of key/value blocks across workers due to forcing workers to transmit KV chunks among neighbors in the same node.   
However, as shown in Fig.~\ref{sfig:hierarchical_comp}, even with the two-level KV transfer (inner and outer ring), computation remains diagonally aligned across steps, preserving the Vertical-Slash pattern and maintaining load balance.

By integrating this hierarchical design into sparse ring attention in \method{}, inter-node KV transfers are fully overlapped with inner-ring computation, effectively mitigating communication overhead from inter-node data movement.

\section{Experiments}
% ----------------------------
In this section, we evaluate \method{} from three perspectives: (i) training efficiency via end-to-end throughput analysis, (ii) training effectiveness via loss convergence, and (iii) downstream long-context performance on RULER, NIAH, InfiniteBench, and PG-19.

\vspace{-5pt}

\begin{figure*}[!tb]
  \vspace{-10pt}
  \centering
  \subfloat[Training Loss Curve.]{
    \label{sfig:training_loss}
    \includegraphics[width=0.38\linewidth]{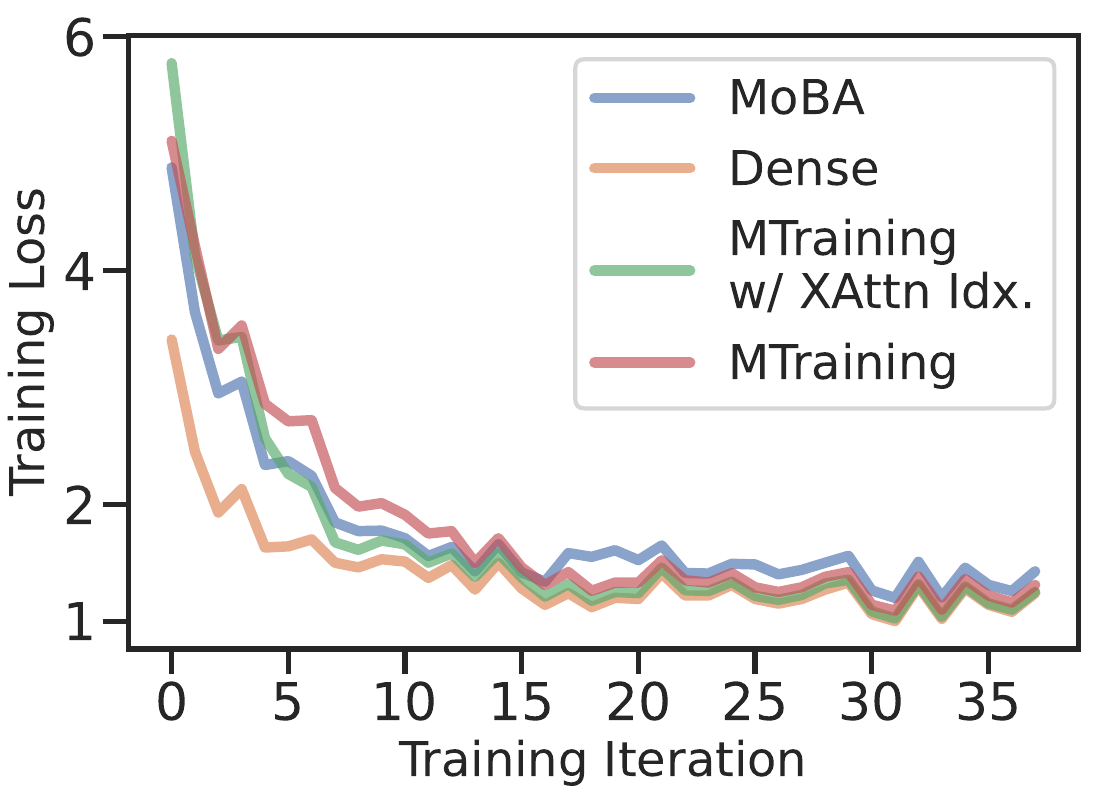}}
  \subfloat[
        Training Throughput Scaling
    ]{
    \label{sfig:training_throughput}
    \includegraphics[width=0.36\linewidth]{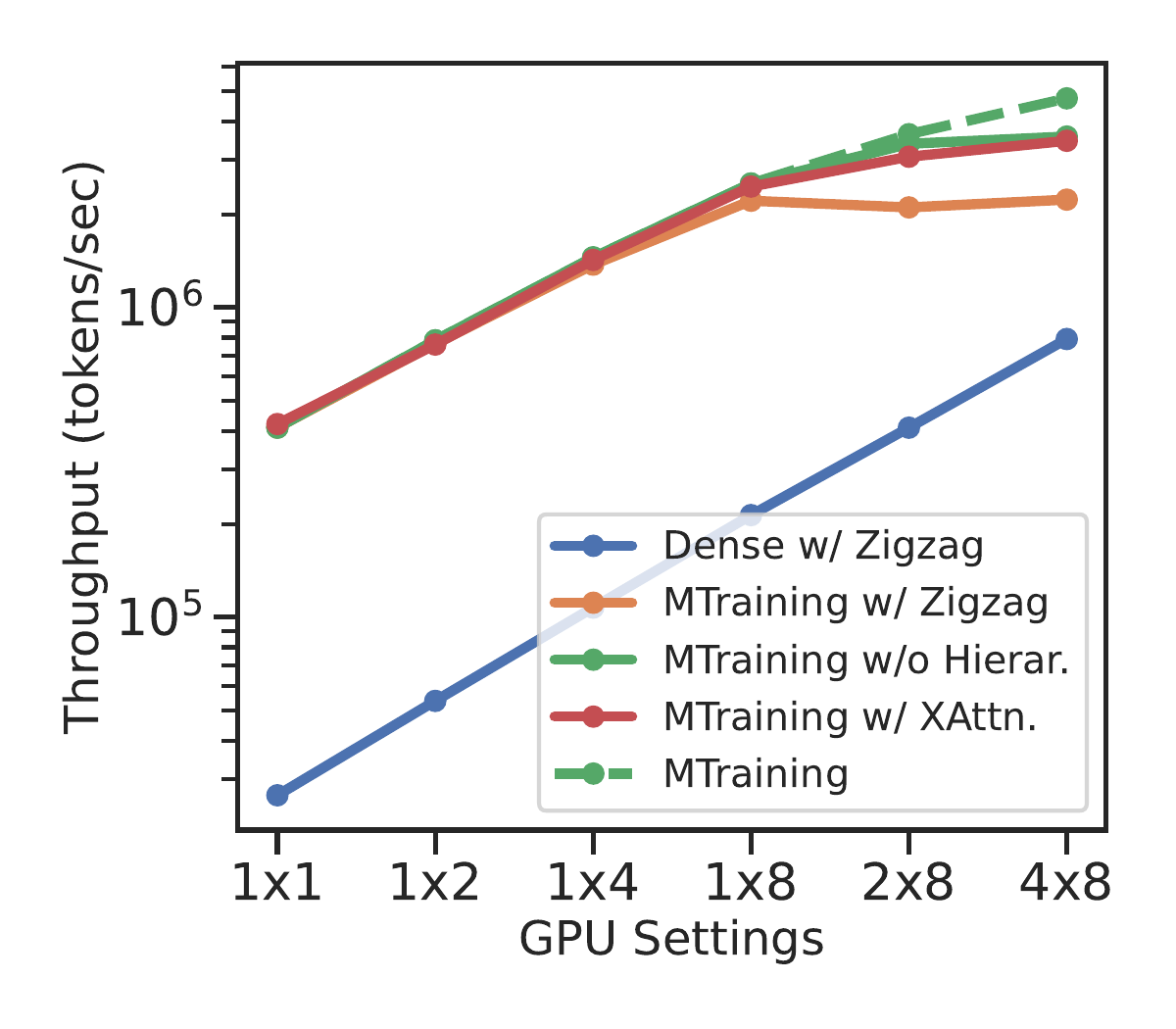}}
    % \hspace{0.2em}
    % \hspace{4em}
  \caption{The training loss and throughput comparison of different methods during continued pretraining of Qwen2.5-3B on the ProLong dataset with a 512K token context window.}
  \label{fig:training}
  % \vspace{-10pt}
\end{figure*}

\subsection{Implementation Details}
\label{subsec:implementation}
% \todo{Double check details}
We conduct experiments using the state-of-the-art open-source LLM Qwen2.5-3B~\cite{yang2024qwen2-5}, trained on a 4×8 NVIDIA A100 40GB cluster. The interconnect includes both InfiniBand and NVLink for high-throughput communication. For attention computation, we employ Context Parallelism = 32. For the remaining components, we use nnScaler~\cite{lin2024nnscaler} to automatically search for the optimal parallelism configuration and keep the same parallel settings across different training runs. To reduce memory consumption, we adopt ZeRO-2 with offloading~\cite{rajbhandari2020zero}, along with gradient accumulation~\cite{huang2019gpipe} and gradient checkpointing~\cite{chen2016training}. All training and inference are performed using the bfloat16 format. We implement a lightweight custom CUDA kernel that builds upon FlashAttention~\cite{dao2022flashattention}, BlockSparse Attention~\cite{guo2024blocksparse}, and the PIT dynamic sparse compiler~\cite{zheng2023pit} to support our method efficiently.
For inference evaluation, we use vLLM~\cite{kwon2023efficient} with greedy decoding to ensure stable and deterministic results across all experiments. Additional experimental details are provided in Appendix~\ref{app:additional_experimental_details}.

% \subsection{Long-context Training}
\subsection{Long-context Training Setup}

% \paragraph{Long-context Extension Training}
\paragraph{Training Settings}

We evaluate our method in the long-context extension stage of training. Specifically, we extend the context window of Qwen2.5-3B from 32K to 512K tokens using Yarn~\cite{peng2023yarn} extrapolated RoPE, with a scaling factor set to 32. Based on this configuration, we perform context extension training on the ProLong dataset~\cite{gao2024prolong}, with a maximum sequence length of 512K tokens, training over 1B tokens for 1 epoch. The averaged sparsity ratio profiled over all data samples with Qwen2.5-3B and \method{} is 0.95. To demonstrate generalizability across model scales and architectures, we additionally train Llama-3.1-8B-Instruct~\cite{grattafiori2024llama3} on ProLong for 2B tokens with 512K context (extending from its original 128K window), following the Final Recipe from~\cite{gao2024prolong}; these results are presented in \S\ref{subsec:additional_exp}.

% \vspace{-16pt}

% \paragraph{Baselines}
\textbf{Baselines} \quad Our baselines serve two complementary purposes for algorithmic training quality and efficiency-oriented throughput evaluation respectively:

\textit{Training quality baselines.} To evaluate the algorithmic effectiveness of \method{}'s sparse attention on model quality, we compare against:
1) \textbf{Dense Attention} without any sparsity, which serves as a reference for convergence behavior; 
2) \textbf{MoBA}~\cite{lu2025moba}, a block-level dynamic sparse attention training method designed for long-context training, whose open-source implementation we adapt to run with Zigzag ring attention;
3) \textbf{\method{} w/ XAttn Idx.}. We replace our online sparse index with XAttention's~\cite{xu2025xattention}, which adopts anti-diagonal pattern to derive the block sparse mask.
At the time of our experiments, MoBA was the only publicly available dynamic sparse attention method with open-source training support specifically designed for long-context training, making it a natural baseline.
XAttention was originally proposed as a \emph{training-free} sparse attention method for inference acceleration; we adapted it as a sparse index builder under hierarchical and balanced ring attention to compare the effectiveness of different sparse patterns.

\textit{Efficiency baselines.} To evaluate the efficiency improvements contributed by each component of our approach, we primarily compare \method{} against 1) \textbf{\emph{Dense attention w/ Zigzag}} and variants of our method with certain components removed:
2) \textbf{\emph{\method{} w/ ZigZag}}, using ZigZag ring attention;
3) \textbf{\emph{\method{} w/o Hierarchical}}, using Striped ring attention without the hierarchical communication scheme.
For comprehensiveness, we also include 4) \textbf{\emph{\method{} w/ XAttn Idx.}} in the throughput evaluation to validate the advantages of our algorithm--system co-design, where the hierarchical and balanced ring attention components are retained but with XAttention's sparse index.

% \subsection{Efficiency Results}

\subsection{End-to-End Throughput Results}
% Fig.~\ref{sfig:training_throughput} illustrate the forward and backward latency, as well as training throughput, of different methods under varying context lengths, distributed worker counts, and sparsity ratios. Notably, \method{} achieves up to 10× end-to-end training speedup at a 512K context length. Compared to Ours w/ ZigZag, and Ours w/o Hierarchical, our method is respectively × and × faster.

Fig.~\ref{sfig:training_throughput} illustrates the training throughput of different methods under distributed worker counts, where the token length is 512K and we manually fix the sparse ratio to be 95\%. Notably, \method{} achieves up to 6× end-to-end training speedup at a 512K context length. Compared to Ours w/ ZigZag, and Ours w/o Hierarchical, our method is respectively 2.1× and 1.3× faster. Moreover, \method{} achieves \textbf{near-linear throughput scaling} with increasing worker count, enabling scalable dynamic sparse attention. In contrast, baseline methods degrade significantly in distributed settings, yielding speedups well below their theoretical limits.

The 2.1$\times$ speedup of \method{} over ``Ours w/ ZigZag'' directly reflects the benefit of workload balancing: ZigZag's mixed diagonal/anti-diagonal computation schedule leads to severe worker-level imbalance (imbalance degree $>$2.4, see \S\ref{subsec:balance_analysis}) that causes faster workers to idle at synchronization barriers. The additional 1.3$\times$ gain over ``Ours w/o Hierarchical'' stems from overlapping inter-node communication with intra-node computation; without the hierarchical design, inter-node transfers (0.98\,ms per step) dominate per-step latency even when computation is reduced by sparsity (see \S\ref{subsec:comm_overlap} for a detailed breakdown).

Notably, even with the hierarchical and balanced ring attention infrastructure retained, \method{} w/ XAttn Idx.\ exhibits worse scaling than ``Ours w/o Hierarchical,'' indicating that the sparse index algorithm itself affects system-level efficiency. It validates the effectiveness of our algorithm-system co-design: the online VS-pattern-aligned index approximation is crucial for maintaining balanced workloads that enable effective communication--computation overlap. 
In contrast, XAttention's anti-diagonal scoring does not provide a good fit for the per-step attention computation distribution in Ring Attention, resulting in suboptimal throughputs in distributed scenarios.

\textbf{Multi-node scalability.} As worker count increases from 8 (single node) to 32 (4 nodes), \method{} maintains its scaling advantage by absorbing the additional inter-node communication cost through its hierarchical design. The near-linear scaling behavior demonstrates that the hierarchical design is essentially effective when the communication fabric becomes heterogeneous (NVLink within nodes vs.\ InfiniBand across nodes).

% % ----------------------------

\subsection{Workload Balance Analysis}
\label{subsec:balance_analysis}

\begin{figure*}[tb]
  % \vspace{-10pt}
  \centering
  \subfloat[
    Worker-level Workload.
    % Workload Distribution over GPUs
    ]{
    \label{sfig:app_worker_imbalance}
    \includegraphics[width=0.9\linewidth]{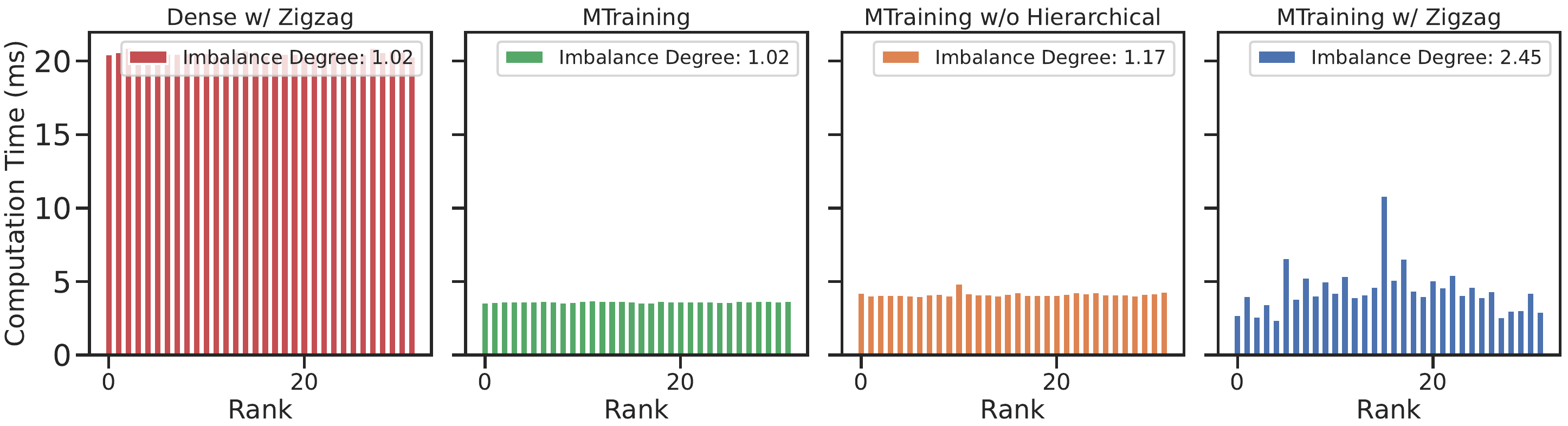}}\\
  \subfloat[Step-level Workload.]{
    \label{sfig:app_step_imbalance}
    \includegraphics[width=0.9\linewidth]{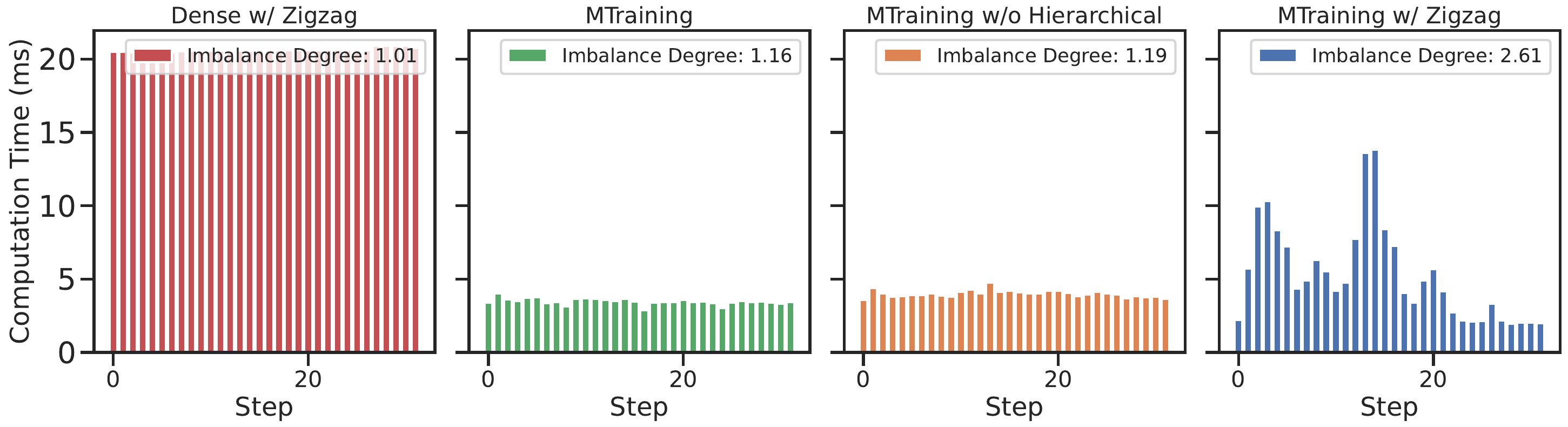}}
    % \hspace{0.2em}
    % \hspace{4em}
  \caption{Distribution of attention computation time using different methods with 512K tokens on 32 GPUs: across CP workers within a fixed Ring Attention step (a) and across Ring Attention steps for a fixed worker (b).}
  \label{fig:imbalanced_additional}
  % \vspace{-10pt}
\end{figure*}

To provide a straightforward illustration on how balanced the worker- and step-level workload become under different training strategies, we measure both worker-level and step-level imbalance during attention computation by imbalance degree ($max/mean$). For the worker-level analysis (Fig. \ref{sfig:app_worker_imbalance}), we randomly sample a Ring Attention step from a randomly selected training iteration and record the computation time of that step across all context-parallel (CP) workers. For the step-level analysis (Fig. \ref{sfig:app_step_imbalance}), we fix one CP worker in the same iteration and measure its attention computation time across all Ring Attention steps. Additionally, Table \ref{tab:imbalance_metrics} in Appendix \ref{app:measure_imbalance} further aggregates statistics across all training iterations, including (i) the average worker-level imbalance degree over all steps and workers, (ii) the average step-level imbalance degree over all workers, and (iii) the average computation time ratio relative to the Ring Attention step latency.

As shown in the results, \method~achieves nearly uniform computation times across both workers and steps, with imbalance degrees close to 1.0, indicating both well-balanced distributed workloads and significantly reduced attention computation time.  Introducing the hierarchical design slightly increases variance but still maintains strong balance. In contrast, applying Zigzag Ring Attention exhibits severe imbalance at both levels (imbalance degree > 2.4), causing inefficient GPU utilization. Overall, these results confirm that \method~effectively mitigates worker- and step-level imbalance in distributed dynamic sparse attention, while achieving over 80\% per-step computation time ratio.

\begin{figure}[!tp]
    % \vspace{-2pt}
    \centering
    \includegraphics[width=1\linewidth]{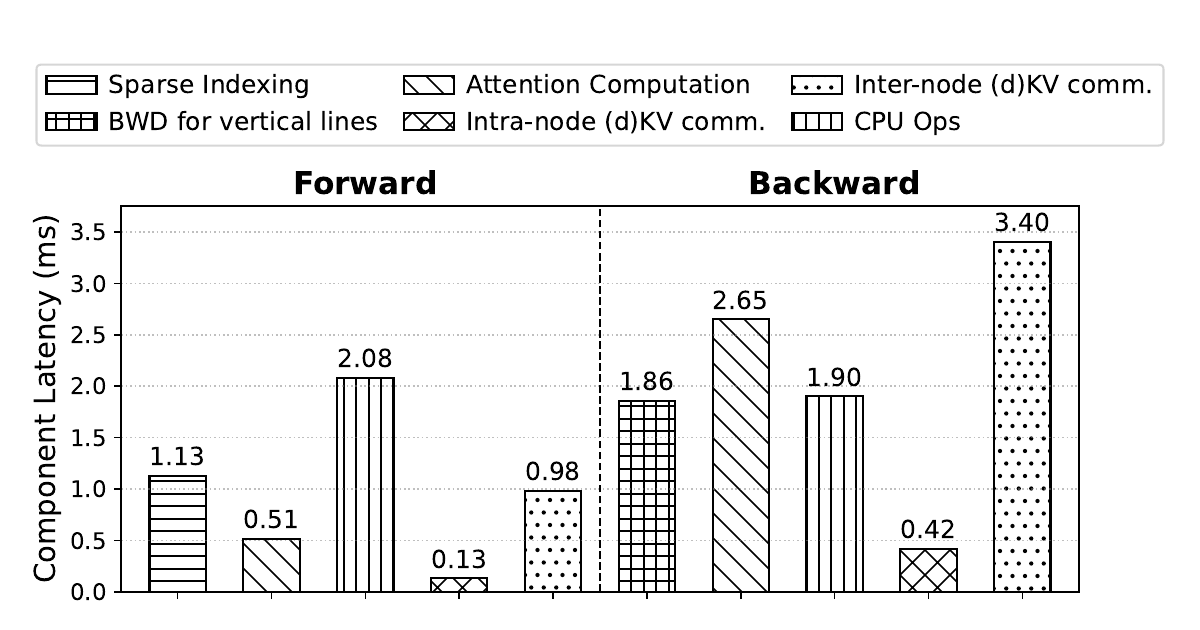}
    \caption{Latency breakdown of forward and backward computation within one Ring Attention step.}
    \label{fig:lat_break}
    \vspace{-10pt}
\end{figure}

\subsection{Computation-Communication Overlap Analysis}
\label{subsec:comm_overlap}

To illustrate how the hierarchical design benefits the distributed long-context training,  we provide a more detailed latency breakdown for a single time of attention computation in training Qwen2.5-3B on a 4-node setup in Fig. \ref{fig:lat_break}, along with the following analysis for the \textbf{forward} process:

Formally, let $T_{\text{comp}}=0.51ms$ be inner-ring compute time, $T_{\text{intra}}=0.13ms$ the intra-node (NVLink) communication time, and $T_{\text{inter}}=0.98ms$ the inter-node (InfiniBand) communication time.

\textbf{Without} the hierarchical design, the communication time of each Ring Attention step takes:
\vspace{-5pt}
\begin{equation*}
    T_{step}\approx max\{T_{comp},T_{intra},T_{inter}\} = T_{\text{inter}}=0.98ms
\end{equation*}
\vspace{-10pt }

Given the communication happens $32-1=31$ times, the total time including sparse index building ($1.13$ ms), CPU operations ($2.08$ ms) and the last time of attention computation ($0.51$ ms) reaches $1.13 + 2.08 + \mathbf{0.98 \times 31} + 0.51 = 34.10$ ms.
% $34.10$ ms.

% \input{tabs/latency_breakdown} 

By effectively overlapping the inter-node communication and inner Ring Attention, \textbf{Hierarchical} Balanced Sparse Ring Attention makes the latency of each step determined by:
\vspace{-5pt}
\begin{equation*}
    T^{hier}_{step}\approx max\{T_{comp}, T_{inner}\} = T_{comp}=0.51ms 
\end{equation*}
\vspace{-5pt}

which happens $32$ times. Taking sparse index building and CPU operation time together, the total time achieves $1.13 + 2.08 + \mathbf{0.51 \times 32} = 19.53$ms, cutting \textbf{42.7\%} of the forward attention time. Similar analysis can be applied to the backward process. Along with Figure \ref{fig:training}, this confirms the effectiveness of our approach in minimizing end-to-end training time.

\vspace{-5pt}
\subsection{Training Loss Analysis}
% \paragraph{Training Loss}

As shown in Fig.~\ref{sfig:training_loss}, we observe the following trends:
1) In the early training stages, dense attention achieves faster loss reduction compared to all sparse methods, which is expected: the sparse approximation introduces an initial gap as the model adapts to attending over a dynamically selected subset of context positions under limited continued pretraining. Importantly, this gap narrows steadily as training proceeds, and \method{}'s loss closely approaches that of dense attention in later stages, demonstrating \emph{convergence} rather than divergence. The model trained with sparse attention does not drift away from the dense baseline.
2) MoBA exhibits a faster initial decline in training loss compared to our method, but its performance deteriorates in later steps, resulting in a significantly higher final training loss than both our method and dense attention. We attribute this divergence to the mismatch between MoBA's coarse-grained, fixed block-level sparsity index and the fine-grained, dynamically shifting attention activations: as training progresses and the model's attention patterns evolve, a rigid block-level selection increasingly misses important attention entries, degrading representational fidelity. In contrast, \method{}'s online top-p budget mechanism adapts continuously to the model's evolving attention distribution, enabling it to maintain close alignment with dense training throughout.

% \textbf{TODO}: Launch 1M-long training with MTraining (not caring about the parallelization strategy)
% \begin{itemize}[leftmargin=2em]
%     \item Model: Qwen-2.5 (7B); 
%     \item Data: 1M dataset (\textbf{TODO})
%     \item Environment: 8xMI300
% \end{itemize}

\subsection{Long-context Downstream Tasks}

\paragraph{Benchmark and Metrics}

To further validate that training with \method{} causes minimal loss to the trained model, we adopt the following benchmarks and metrics to evaluate the effectiveness of \method{}: 1) \textbf{RULER}~\cite{hsieh2024ruler}, a comprehensive benchmark comprising 13 tasks across 4 categories, including retrieval, multi-hop reasoning, information aggregation, and question answering. 2) \textbf{Needle In A Haystack (NIAH)} \cite{kamradt2023needle} assesses LLMs' ability to retrieve key information placed at various positions in a long context. 3) \textbf{PG19} \cite{rae2019pg19}, a long-form language modeling benchmark with sequences up to 300K tokens. We report perplexity to measure language modeling performance. 4) \textbf{InfiniteBench}~\cite{zhang2024InfiniteBench} is a comprehensive benchmark for long-context language processing, including question answering, code debugging, summarization etc., with average context length of 214K tokens.

\begin{table}[tp]
    \caption{Performance (\%) of various training–inference combinations on RULER~\cite{hsieh2024ruler} at context lengths from 16K to 512K with the long-context-extended Qwen2.5-3B.}
    % where \emph{MTraining w/ XAttn Idx.} indicate XAttention is applied for computing the block sparse index, and \emph{Inference} indicates the attention algorithm used for inference during evaluation.}
    \centering
    \small
    \setlength{\tabcolsep}{4pt}      % horizontal padding
    \renewcommand{\arraystretch}{1.1}% vertical padding
    % \resizebox{0.84\columnwidth}{!}{
    % \begin{tabular}{l l|cccccc|c}
    \resizebox{\columnwidth}{!}{
    \begin{tabular}{l l|cccccc|c}
        \toprule
        \textbf{Training} & \textbf{Inference} & \textbf{16K} & \textbf{32K} & \textbf{64K} & \textbf{128K} & \textbf{256K} & \textbf{512K} & \textbf{Avg.} \\
        \midrule
        Dense                        & Dense      & 72.51 & 67.83 & 58.46 & 52.38 & 55.91 & 54.15 & 60.21 \\
        Dense                        & MInference & 54.58 & 54.97 & 49.85 & 43.93 & 38.83 & 41.10 & 47.21 \\
        MoBA                         & Dense      & 64.61   & 55.06   & 45.44   & 38.24   & 35.48   & 34.99   & 45.64   \\
        {\cellcolor[rgb]{0.925,0.957,1}}\textbf{MTraining} w/ XAttn Idx. & {\cellcolor[rgb]{0.925,0.957,1}}Dense      & {\cellcolor[rgb]{0.925,0.957,1}}75.04 & {\cellcolor[rgb]{0.925,0.957,1}}67.97 & {\cellcolor[rgb]{0.925,0.957,1}}58.93 & {\cellcolor[rgb]{0.925,0.957,1}}51.43 & {\cellcolor[rgb]{0.925,0.957,1}}56.96 & {\cellcolor[rgb]{0.925,0.957,1}}54.85 & {\cellcolor[rgb]{0.925,0.957,1}}60.86 \\
        {\cellcolor[rgb]{0.925,0.957,1}}\textbf{MTraining}                    & {\cellcolor[rgb]{0.925,0.957,1}}Dense      & {\cellcolor[rgb]{0.925,0.957,1}}\textbf{76.13} & {\cellcolor[rgb]{0.925,0.957,1}}\textbf{70.51} & {\cellcolor[rgb]{0.925,0.957,1}}60.81 & {\cellcolor[rgb]{0.925,0.957,1}}\textbf{58.65} & {\cellcolor[rgb]{0.925,0.957,1}}\textbf{58.33} & {\cellcolor[rgb]{0.925,0.957,1}}\textbf{54.88} & {\cellcolor[rgb]{0.925,0.957,1}}\textbf{63.22} \\
        {\cellcolor[rgb]{0.925,0.957,1}}\textbf{MTraining}                    & {\cellcolor[rgb]{0.925,0.957,1}}MInference & {\cellcolor[rgb]{0.925,0.957,1}}75.44 & {\cellcolor[rgb]{0.925,0.957,1}}69.60 & {\cellcolor[rgb]{0.925,0.957,1}}\textbf{62.92} & {\cellcolor[rgb]{0.925,0.957,1}}53.19 & {\cellcolor[rgb]{0.925,0.957,1}}51.59 & {\cellcolor[rgb]{0.925,0.957,1}}50.85 & {\cellcolor[rgb]{0.925,0.957,1}}60.60 \\
        \bottomrule
    \end{tabular}
    }
    \label{tab:ruler_results_filled}
\end{table}

% 64.61,55.06,45.44,38.24,35.48,34.99

\begin{figure}[t!]
    % \vspace{-2pt}
    \centering
    \includegraphics[width=0.9\linewidth]{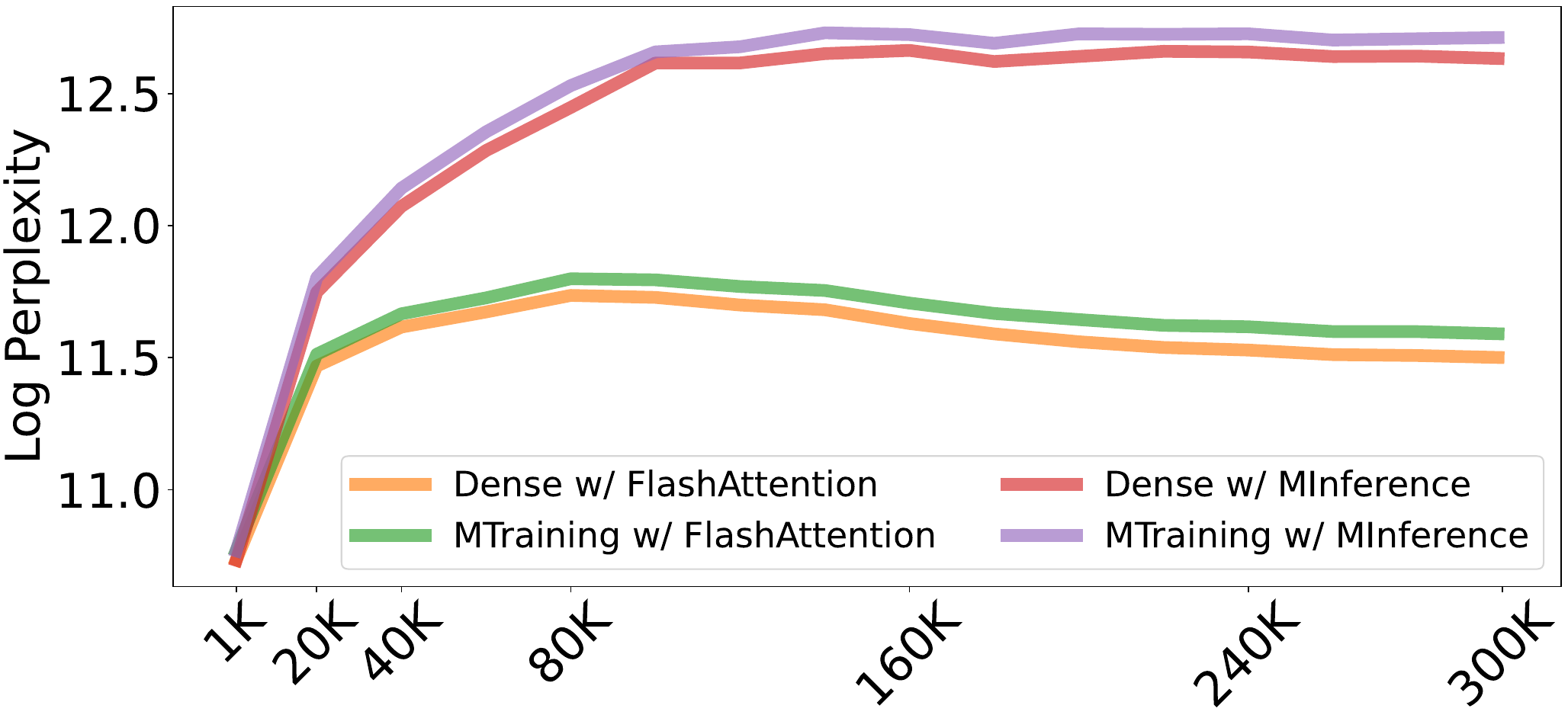}
    \caption{Language Modeling Results on PG19.}
    \label{fig:pg19_results}
    \vspace{-10pt}
\end{figure}

\paragraph{RULER}
To further evaluate long-context capability, we benchmark \method{} on RULER, a state-of-the-art long-context evaluation suite. As shown in Table~\ref{tab:ruler_results_filled}, \method{} consistently outperforms baselines across various context lengths.
Compared to dense training, \method{} achieves 3\% and 13.4\% overall improvement under dense and MInference-based inference, respectively—reaching up to 6.3\% gain at 128K tokens. Additionally, \method{} outperforms its variant with fixed XAttn indexing by 2.4\%, highlighting that training-time dynamics affect the representativeness of anti-diagonal structures.

\vspace{-5pt}
\paragraph{Needle In A Haystack}
% \textbf{Needle In A Haystack}
% 

\begin{figure*}[htb]
    \centering
    % \vspace{-12pt}
    \subfloat[Full Attention]{
      \label{sfig:niah_baseline}
      \includegraphics[width=\columnwidth]{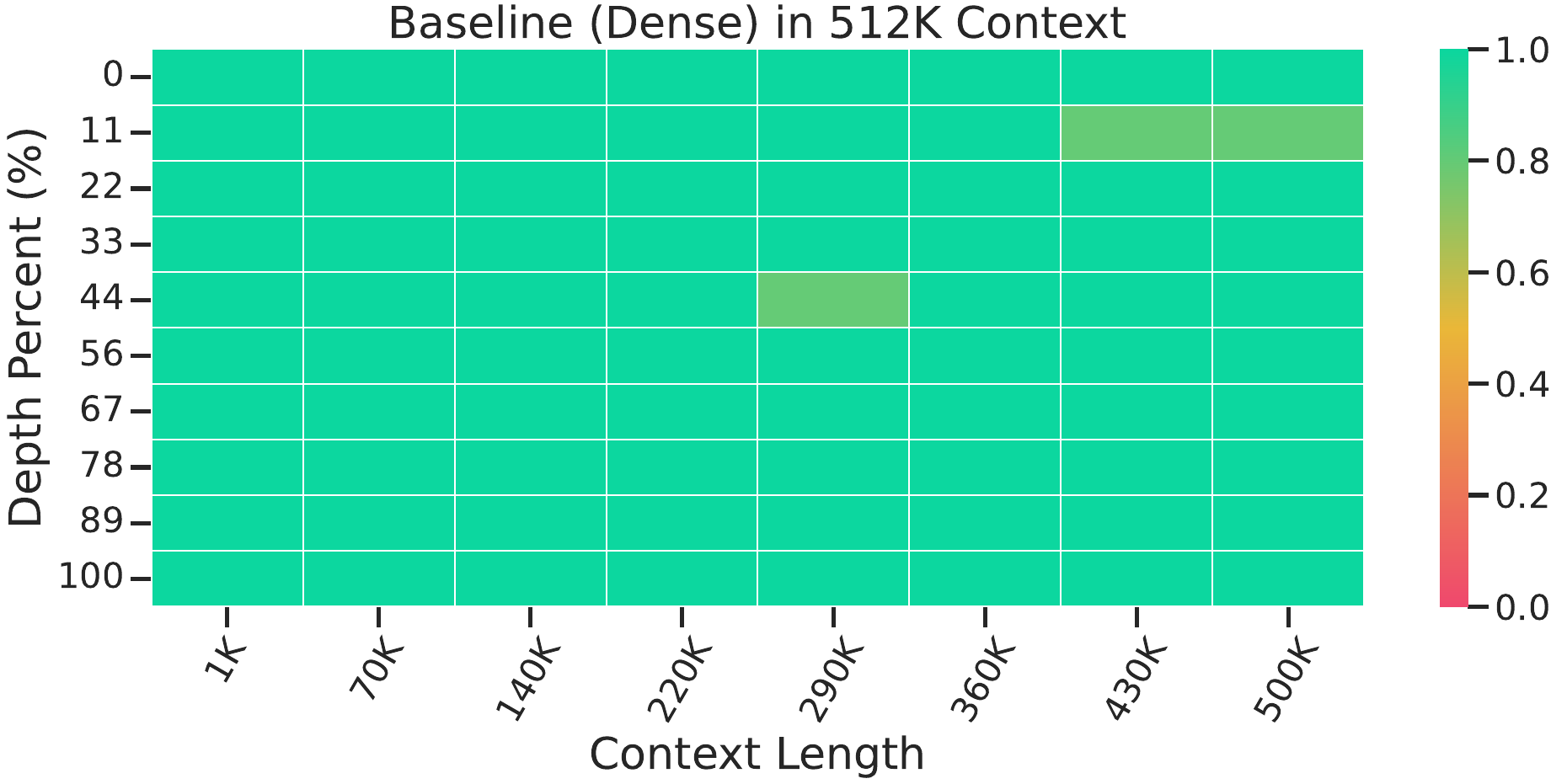}}
    \subfloat[MTraining]{
      \label{sfig:niah_mtraining}
      \includegraphics[width=\columnwidth]{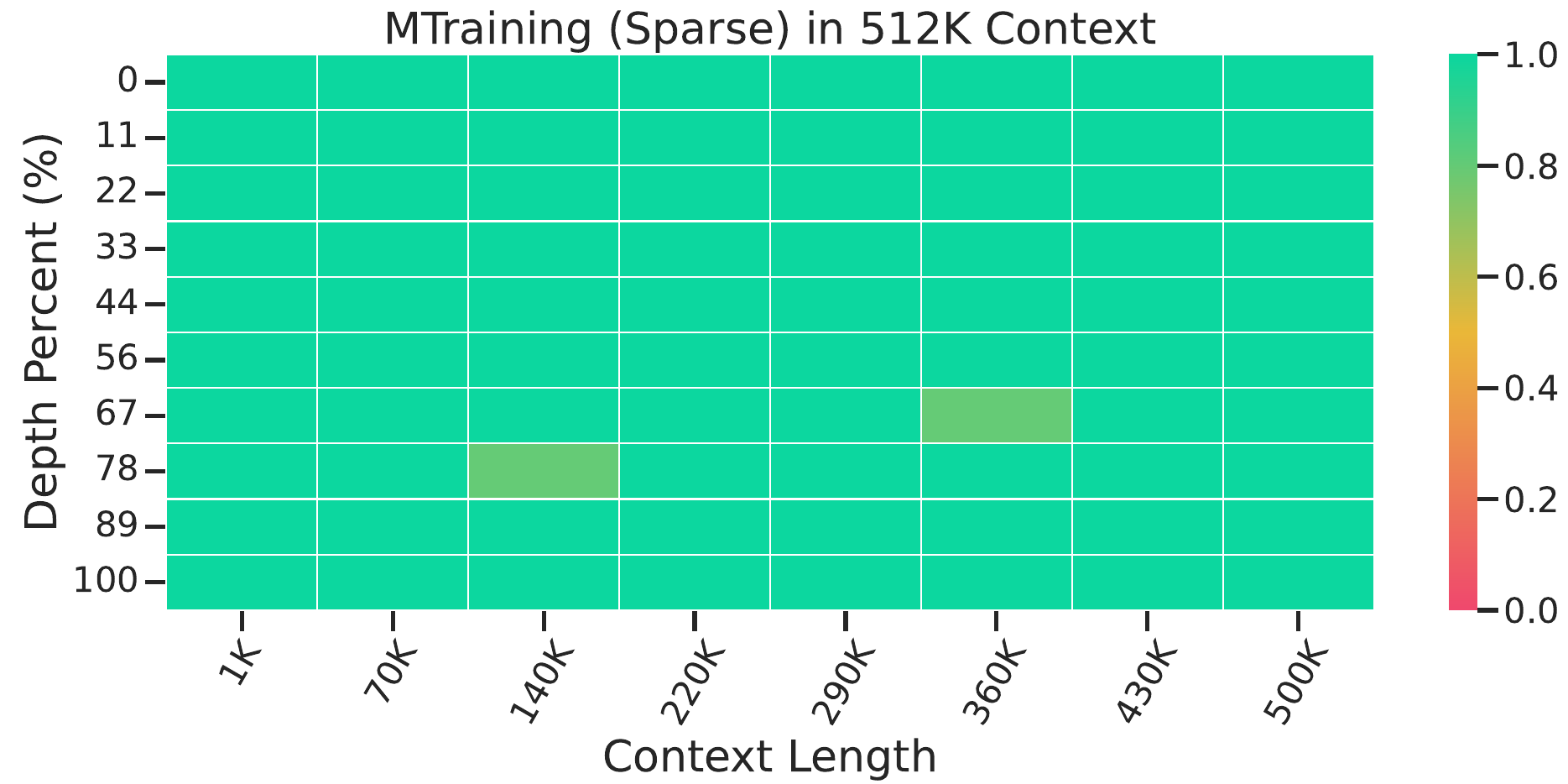}}
    \caption{Needle In A Haystack Results of the baseline checkpoint and the MTraining checkpoint.}
    \label{fig:niah_results}
    \vspace{-8pt}
\end{figure*}

As shown in Fig.~\ref{fig:niah_results}, MTraining achieves near-perfect retrieval performance on the NIAH. Compared to the baseline, MTraining yields better overall retrieval accuracy, despite MTraining's largely reduced computational cost. We also report the NIAH results of the baseline and the MTraining checkpoint w/ MInference in the inference stage shown in Fig.~\ref{fig:niah_results_minference}, where MTraining w/ MInference also achieves a better overall retrieval accuracy than the baseline checkpoint.

% \begin{wrapfigure}{r}{\columnwidth}
%     % \vspace{-20pt}
%     \centering
%     \includegraphics[width=1\linewidth]{figs/pg19.pdf}
%     \caption{Language Modeling Results on PG19.}
%     \label{fig:pg19_results}
%     % \vspace{-20pt}
% \end{wrapfigure}
% \begin{figure}[t!]
%     % \vspace{-2pt}
%     \centering
%     \includegraphics[width=0.9\linewidth]{figs/pg19.pdf}
%     \caption{Language Modeling Results on PG19.}
%     \label{fig:pg19_results}
%     \vspace{-10pt}
% \end{figure}

\vspace{-5pt}
\paragraph{Language Modeling}
% \textbf{Language Modeling}

We evaluate the language modeling performance of MTraining against the baselines on the PG19 dataset with perplexity as the metric.
As shown in Fig.~\ref{fig:pg19_results}, \method{} maintains perplexity that closely tracks the dense baseline across all evaluated context lengths, with the relative difference remaining within a small margin. 
The consistent \emph{tracking} behavior confirms that the sparse approximation does not introduce systematic modeling errors. The same trend holds when MInference is applied at inference time.

\vspace{-5pt}
\paragraph{InfiniteBench}
% \textbf{InfiniteBench}
% \begin{wraptable}{r}{0.5\textwidth}
%     \small
%     \centering
%     \setlength{\tabcolsep}{0.5mm}
%     \vspace{-3ex}
%     \caption{Performance (\%) on InfiniteBench~\cite{zhang2024InfiniteBench}.}
%     \resizebox{0.5\columnwidth}{!}{
%     \begin{tabular}{lcccccc}
%     \toprule
%         Methods &  \textbf{En.Sum} & \textbf{En.QA} & \textbf{En.MC} & \textbf{En.Dia} & \textbf{Code.Debug} &  \textbf{Avg.} \\
%         \midrule
%         Dense & 18.5 & \textbf{8.2} & 63.5 & \textbf{6.0} & 26.3 & 24.5 \\
%          w/ MInference & 17.4 & 6.7 & 63.0 & 4.4 & 17.0 & 21.7 \\
%         \midrule
%         MTraining & 19.5 & 6.8 & \textbf{65.3} & 3.5 & \textbf{34.1} & \textbf{25.8} \\
%          w/ MInference & \textbf{19.5} & 6.7 & 63.3 & 5.0 & 15.5 & \textbf{22.0} \\
%     \bottomrule
%     \end{tabular}
%     }
%     \vspace{-2ex}
%     \label{tab:infinitebench}
% \end{wraptable}

\vspace{-10pt}
\begin{table}[t]
    \small
    \centering
    \setlength{\tabcolsep}{0.5mm}
    \caption{Performance (\%) on InfiniteBench~\cite{zhang2024InfiniteBench}.}
    \resizebox{0.5\textwidth}{!}{
    \begin{tabular}{lcccccc}
    \toprule
        Methods &  \textbf{En.Sum} & \textbf{En.QA} & \textbf{En.MC} & \textbf{En.Dia} & \textbf{Code.Debug} &  \textbf{Avg.} \\
    \midrule
        Dense & 18.5 & \textbf{8.2} & 63.5 & \textbf{6.0} & 26.3 & 24.5 \\
         w/ MInference & 17.4 & 6.7 & 63.0 & 4.4 & 17.0 & 21.7 \\
    \midrule
        MTraining & 19.5 & 6.8 & \textbf{65.3} & 3.5 & \textbf{34.1} & \textbf{25.8} \\
         w/ MInference & \textbf{19.5} & 6.7 & 63.3 & 5.0 & 15.5 & \textbf{22.0} \\
    \bottomrule
    \end{tabular}
    }
    \label{tab:infinitebench}
    \vspace{-10pt}
\end{table}

As shown in Table~\ref{tab:infinitebench}, \method{} achieves superior performance on InfiniteBench compared to the dense baseline. Specifically, \method{} improves the coding and summarization capabilities compared to the baseline, while maintaining a competitive performance on the question answering tasks. We also report the results with MInference in the inference stage, which also shows a similar trend.

\subsection{Additional Experiments}
\label{subsec:additional_exp}

We conduct two additional studies to evaluate the scalability and analyse the sparse ratio of \method{} during training with detailed results and analysis provided in Appendix~\ref{sec:scalability_appendix} and~\ref{app:sparsity_dynamics}.

\textbf{Generalization to larger model scales.} We further train Llama-3.1-8B-Instruct~\cite{grattafiori2024llama3} on ProLong for 2B tokens, and find that \method{} achieves nearly lossless RULER accuracy compared to dense training (68.07 vs.\ 69.07) with substantially better sparse-inference robustness (68.58\% vs.\ 30.60\%).

\textbf{Sparsity dynamics.} By evaluating checkpoints across iterations of Qwen training, we confirm that the global sparsity ratio remains stable at $93.7 \pm 3.9\%$, where layer-level sparsity exhibits a general increasing trend from 87.3\% to 99.6\%, guaranteeing consistent high sparsity for computation acceleration.
\section{Related Work}

\paragraph{Long-context Training System}

To scale long-context LLM training, various parallelization strategies have been developed, including activation parallelism \cite{korthikanti2023megatron-sp}, distributed attention methods \cite{jacobs2023ulysses, liu2024ring-attention}, and offloading-based approaches \cite{luo2024minisequence}. Among them, Ring Attention \cite{liu2024ring-attention} offers the best scalability by distributing KV computation via P2P communication with block-wise computation. However, it still faces two key challenges: communication overhead and worker imbalance.
Variants such as Striped \cite{brandon2023stripe} and Zigzag Ring Attention \cite{zhuzilin2024zigzag} address imbalance, while hybrid systems~\cite{fang2024usp, gu2024loongtrain} combine the benefits of Ring Attention and Ulysses. More recent work~\cite{ge2025bytescale, wang2025wlb, wang2025flexsp} improves scheduling for heterogeneous sequence lengths induced by sequence packing\cite{krell2021seq-packing}, and Magi-Attention\cite{magiattention2025} further boosts efficiency through fused kernels and overlapped communication.
Despite these advancements, we emphasize that \method{} is \emph{orthogonal} to these dense-context system optimizations: while existing work accelerates dense attention computation and communication scheduling, our contribution addresses the distinct efficiency and load-balancing challenges that arise specifically when dynamic sparse attention is introduced into distributed long-context training. 
Combining \method{} with advances in dense-context systems remains a promising direction for future work.

\vspace{-5pt}
\paragraph{Scaling Context Windows of LLMs}

Several approaches have been proposed to scale the context window of LLMs, which can be broadly categorized into three groups:
1) Staged pretraining~\cite{liu2024deepseek-v3, yang2025qwen1M, grattafiori2024llama3, gao2024prolong}, which trains the model in multiple stages using data of increasing sequence lengths;
2) Positional embedding manipulation, including extrapolation\cite{su2024rope, ding2024longrope, gao2024prolong, sun2022xPOS} and interpolation\cite{chen2023position-interpolation, peng2023yarn}, which adjust positional encodings to extend models’ sensitivity to longer inputs;
3) Length extrapolation~\cite{an2024trainingfree,jin2024llm,an2025why}, where models trained on short sequences are expected to generalize to longer contexts.

\vspace{-5pt}
\paragraph{Efficiency Enhancement for Long-Context LLMs}

For Transformer-based LLMs, extensive research has focused on improving computational and memory efficiency as input lengths increase. These efforts largely fall into two categories:
1) KV cache optimization, including quantization~\cite{liu2024kivi, hooper2024kvquant}, sharing~\cite{sun2024yoco, goldstein2024goldfinch, ainslie2023gqa, chen2024dha}, and offloading~\cite{jin2024kv-offload, lee2024infinigen};
2) Attention efficiency, aimed at mitigating the quadratic cost of self-attention, using static or clustered sparse patterns~\cite{beltagy2020longformer, zaheer2020bigbird, kitaev2020reformer} and dynamic sparse attention~\cite{jiang2024minference,li2025mminference,lai2025flexprefill, tang2024quest, xu2025xattention, ribar2023sparq, zhang2025spargeattn, chen2024magicpig}.
Recent works such as NSA\cite{yuan2025nsa} and MoBA\cite{lu2025moba} leverage dynamic sparse attention during pretraining to achieve significant speedups with near-lossless accuracy compared to dense baselines. However, scaling dynamic sparse attention efficiently in distributed training remains an open problem.
\section{Conclusion and Discussions}

We propose MTraining, a distributed training methodology that leverages dynamic sparse attention for efficient ultra-long-context LLM training. By combining balanced and hierarchical sparse ring attention, MTraining addresses worker- and step-level imbalance in distributed dynamic sparse attention settings. We validate MTraining by extending Qwen2.5-3B and Llama-3.1-8B-Instruct from 32K/128K to 512K context via continued pretraining on ProLong using 32 A100 GPUs, achieving up to 6$\times$ training throughput while maintaining or improving model accuracy on RULER, PG-19, InfiniteBench, and NIAH.
\vspace{-0.8em}
\paragraph{Limitations and Discussion}
\method{}'s theoretical analysis and sparse index design rely on the Vertical-Slash pattern induced by Rotary Position Embeddings (RoPE). While this limits the formal guarantees to RoPE-based models, we note that the vast majority of modern open-source LLMs, such as Qwen, employ RoPE or its variants like YaRN~\cite{peng2023yarn}, making \method{} broadly applicable in practice. More generally, any positional embedding scheme whose attention expectation is primarily governed by relative position may give rise to analogous sparse structures amenable to our approach. Empirical validation on non-RoPE architectures remains valuable future work.

Our experiments are done in continued pretraining for context extension as it is the typical setting for ultra-long-context LLMs. However, due to limited GPU resources, our experiments focus on models up to 8B parameters with up to 2B training tokens. It would be valuable to validate \method{} at large-scale models for longer training runs.
On the hand, exploring alternative design choices such as threshold heuristics and sparsity estimation granularity remain important future directions.

\newpage
\bibliography{reference}
\bibliographystyle{mlsys2025}

%%%%%%%%%%%%%%%%%%%%%%%%%%%%%%%%%%%%%%%%%%%%%%%%%%%%%%%%%%%%%%%%%%%%%%%%%%%%%%%
%%%%%%%%%%%%%%%%%%%%%%%%%%%%%%%%%%%%%%%%%%%%%%%%%%%%%%%%%%%%%%%%%%%%%%%%%%%%%%%
% SUPPLEMENTAL CONTENT AS APPENDIX AFTER REFERENCES
%%%%%%%%%%%%%%%%%%%%%%%%%%%%%%%%%%%%%%%%%%%%%%%%%%%%%%%%%%%%%%%%%%%%%%%%%%%%%%%
%%%%%%%%%%%%%%%%%%%%%%%%%%%%%%%%%%%%%%%%%%%%%%%%%%%%%%%%%%%%%%%%%%%%%%%%%%%%%%%
\appendix
% \appendix

\newpage
\section{Artifact Availability}
\label{sec:artifact}

The artifact accompanying this paper has been archived on Zenodo with a persistent DOI: \href{https://doi.org/10.5281/zenodo.19484894}{10.5281/zenodo.19484894}.

The source code for \method{} is released as part of the open-source MInference library, available on GitHub at \href{https://github.com/microsoft/MInference/tree/main/mtraining}{github.com/microsoft/MInference}. Within the repository, the training logic of \method{} is located under \texttt{mtraining/}, while the implementation of the distributed sparse attention operators resides under \texttt{minference/dist\_ops/}.

% \section{Limitations}
% \label{sec:limitations}

% Due to limited GPU resources, we conduct experiments on up to 8B-level LLMs with 2B training tokens (see Appendix \ref{sec:scalability} for 8B-level experiments), and do not validate \method{} under large-scale continued pretraining. As context length decreases, communication overhead increasingly dominates, diminishing the benefits of distributed training.6

% \section{Broader Impacts}
% \label{sec:impacts}

% \method{} effectively accelerates long-context LLM training, improving training efficiency in scenarios such as agentic, reinforcement learning training, and code understanding—thereby enabling stronger reasoning capabilities under the same computational budget.

\section{Scalability of MTraining}
\label{sec:scalability_appendix}

We extend \method{} to Llama-3.1-8B-Instruct~\cite{grattafiori2024llama3}, which uses grouped-query attention with a 128K native context window. This expands the model size from 3B to 8B parameters, changes the model architecture from Qwen-2.5 to Llama-3, and increases the amount of training tokens from 1B to 2B, providing a comprehensive test of \method{}'s scalability and generalizability. We extend its context to 512K tokens by continued pretraining on ProLong~\cite{gao2024prolong} for 2B tokens. Other settings including model initialization, RoPE, learning rate, and optimizers follow the Final Recipe (Stage~2 of Continued Pretraining) from~\cite{gao2024prolong}.

\begin{figure}[t]
    \centering
    \includegraphics[width=0.9\columnwidth]{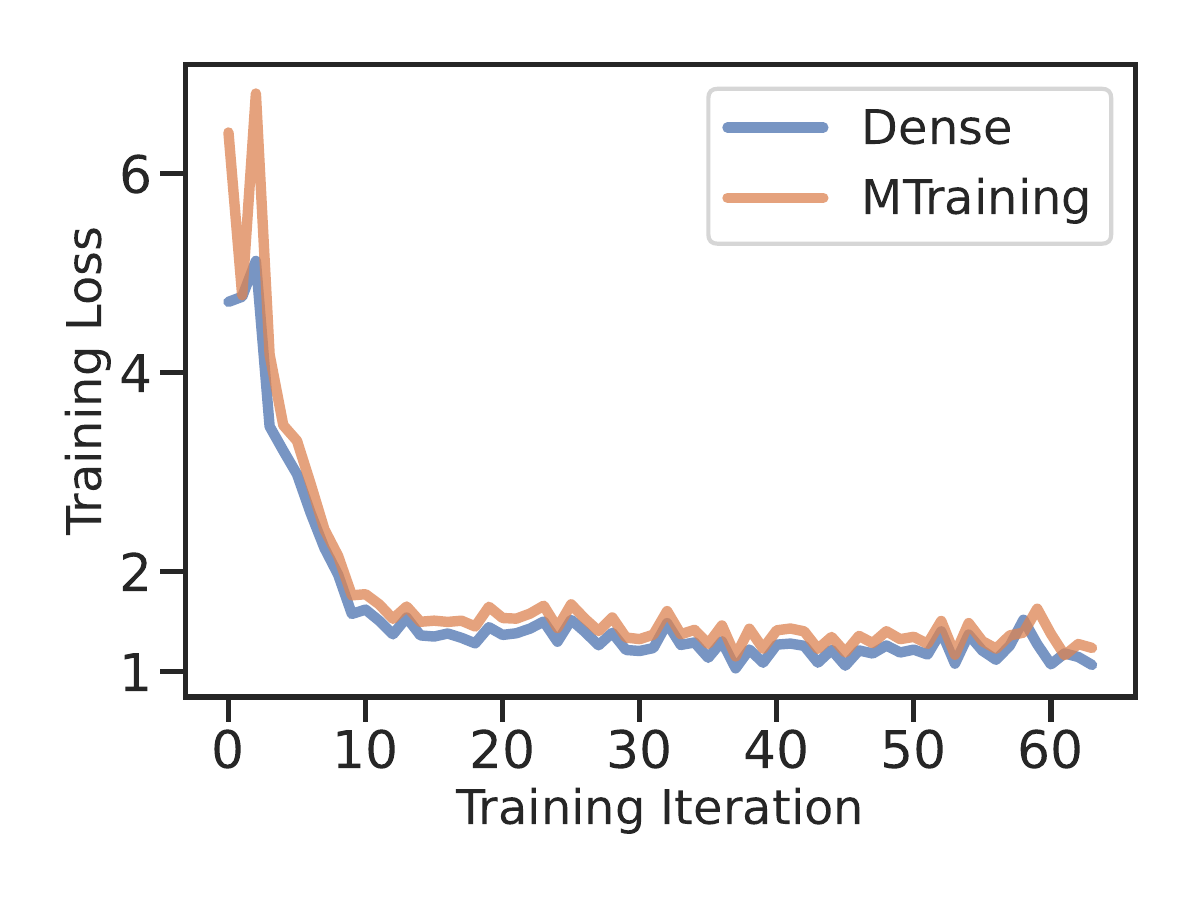}
    \caption{Training loss comparison of dense attention and \method{} during continued pretraining of Llama-3.1-8B-Instruct on ProLong with a 512K-token context window. The sparse model closely tracks the dense baseline throughout training.}
    \label{fig:train-loss-llama}
\end{figure}

\begin{table}[t]
    \caption{Performance (\%) of various training–inference combinations on RULER~\cite{hsieh2024ruler} at context lengths from 16K to 512K with the long-context-extended Llama-3.1-Instruct-8B.}
    \centering
    \setlength{\tabcolsep}{4pt}      % horizontal padding
    \renewcommand{\arraystretch}{1.1}% vertical padding
    \resizebox{\columnwidth}{!}{
    \begin{tabular}{l l|cccccc|c}
        \toprule
        \textbf{Training} & \textbf{Inference} & \textbf{16K} & \textbf{32K} & \textbf{64K} & \textbf{128K} & \textbf{256K} & \textbf{512K} & \textbf{Avg.} \\
        \midrule
        Dense      & Dense      & 82.58 & 80.13 & 73.33 & 62.17 & 59.15 & 57.06 & 69.07 \\
        Dense      & MInference & 55.22 & 45.41 & 25.99 & 22.40 & 19.37 & 15.19 & 30.60 \\
        {\cellcolor[rgb]{0.925,0.957,1}}\textbf{MTraining} & {\cellcolor[rgb]{0.925,0.957,1}}Dense      & {\cellcolor[rgb]{0.925,0.957,1}}83.81 & {\cellcolor[rgb]{0.925,0.957,1}}75.73 & {\cellcolor[rgb]{0.925,0.957,1}}70.15 & {\cellcolor[rgb]{0.925,0.957,1}}64.48 & {\cellcolor[rgb]{0.925,0.957,1}}59.47 & {\cellcolor[rgb]{0.925,0.957,1}}54.80 & {\cellcolor[rgb]{0.925,0.957,1}}68.07 \\
        {\cellcolor[rgb]{0.925,0.957,1}}\textbf{MTraining} & {\cellcolor[rgb]{0.925,0.957,1}}MInference & {\cellcolor[rgb]{0.925,0.957,1}}82.42 & {\cellcolor[rgb]{0.925,0.957,1}}77.10 & {\cellcolor[rgb]{0.925,0.957,1}}70.67 & {\cellcolor[rgb]{0.925,0.957,1}}65.56 & {\cellcolor[rgb]{0.925,0.957,1}}61.13 & {\cellcolor[rgb]{0.925,0.957,1}}54.58 & {\cellcolor[rgb]{0.925,0.957,1}}\textbf{68.58} \\
        \bottomrule
    \end{tabular}
    }
    \label{tab:ruler_llama}
    \vspace{-5pt}
\end{table}

As shown in Fig.~\ref{fig:train-loss-llama}, \method{} exhibits only a minimal training loss gap compared to dense attention throughout the entire training process, maintaining the same convergence trend when applied to the 8B-scale model.
Table~\ref{tab:ruler_llama} presents the RULER evaluation results: \method{} achieves nearly lossless performance compared to dense training (68.07 vs.\ 69.07 average accuracy with dense inference). Notably, \method{} demonstrates better robustness under sparse inference---when MInference is applied at test time, \method{}'s accuracy (68.58\%) substantially exceeds that of the dense-trained model (30.60\%), suggesting that training with dynamic sparse attention produces representations that are inherently more amenable to sparse inference.

\section{Sparsity Dynamics Analysis}
\label{app:sparsity_dynamics}

\begin{table}[t]
    % \small
    \vspace{-10pt}
    \centering
    \caption{Average sparsity ratio (\%) across training iterations and samples for checkpoints of Qwen2.5-3B at 512K context, where 'Global' denotes the overall sparsity ratio across all layers, and 'L' denotes the sparse ratio of each layer.}
    \setlength{\tabcolsep}{3.5pt}
    \renewcommand{\arraystretch}{1.1}
    \resizebox{\columnwidth}{!}{
    \begin{tabular}{l|ccccccc}
        \toprule
        \textbf{Global} & \textbf{L0} & \textbf{L8} & \textbf{L16} & \textbf{L24} & \textbf{L32} & \textbf{L35} \\
        \midrule
        $93.7{\scriptstyle\pm3.9}$ & $87.3{\scriptstyle\pm0.6}$ & $90.8{\scriptstyle\pm1.9}$ & $96.7{\scriptstyle\pm0.8}$ & $95.6{\scriptstyle\pm0.6}$ & $99.6{\scriptstyle\pm0.1}$ & $97.1{\scriptstyle\pm0.3}$ \\
        \bottomrule
    \end{tabular}
    }
    \label{tab:sparsity_ratios}
    \vspace{-10pt}
\end{table}

A natural question is whether the dynamic sparsity ratios used by \method{} remain stable throughout training or exhibit large fluctuations that could undermine workload balance.
To investigate, we loaded checkpoints saved at different iterations during training Qwen2.5-3B and collected the sparsity ratio of each attention layer by performing model inference on a randomly selected set of 512K-token samples from ProLong.
The results are summarized in Table~\ref{tab:sparsity_ratios}.

The global sparsity ratio remains stable at $93.7 \pm 3.9\%$ throughout training, fluctuating within the 90--98\% range without a systematic increasing or decreasing trend. This stability is a consequence of the top-p mass threshold mechanism, which automatically adapts the sparsity budget per attention head based on the current attention distribution rather than imposing a fixed top-k budget.

Along the layer dimension, sparsity generally exhibits an increasing trend from lower to higher layers, where Layer~32 reaches $99.6 \pm 0.1\%$ with a very small standard deviation across training iterations and samples. Notably, the per-layer ratios are also stable across training iterations as indicated by small standard deviations, indicating that the overall sparsity structure does not drift during training. 

\section{Proof of Theory}
\label{appendix:proof}

% \subsection{Vetrical-Slash Pattern in RoPE}
\subsection{The Gradient of Attention}

\begin{equation}
\begin{aligned}
    % dV &= A dO\\ 
    \frac{\partial \mathcal{L}}{\partial V} &= A^\top \cdot \frac{\partial \mathcal{L}}{\partial \text{O}},\\
    \frac{\partial \mathcal{L}}{\partial Q} &= \frac{1}{\sqrt{d}} \cdot \frac{\partial \mathcal{L}}{\partial S} \cdot K, \\
    \frac{\partial \mathcal{L}}{\partial K} &= \frac{1}{\sqrt{d}} \cdot \left(\frac{\partial \mathcal{L}}{\partial S}\right)^\top \cdot Q\\
\end{aligned}
\label{eq:attention_backward}
\end{equation}

\subsection{Theorem 3.1}

% \todo{Slash proof}

Let $\vec{q}_n \in \mathbb{R}^{1\times d}$ and $\vec{k}_m \in \mathbb{R}^{1\times d}$, where $n, m \in [0, N)$, be the query and key vectors before applying RoPE, respectively. After applying RoPE, their dot product $z_{n,m}$ is calculated as follows:
\begin{equation}
    \begin{aligned}
    \label{eq:z}
    z_{n,m} &= \mathrm{RoPE}(\vec{q}_n, n) ~ \mathrm{RoPE}(\vec{k}_m, m)^T \\
    &= \vec{q}_n \vec{W}_n \vec{W}_m^T \vec{k}_m^T \\
    &= \vec{q}_n \vec{W}_{n-m} \vec{k}_m^T,
    \end{aligned}
\end{equation}

According to the definition of rotary matrices, the dot product $z_{n,m}$ can be further simplified as follows:
\begin{equation}
    \begin{aligned}
        \label{eq:z_simplified}
        z_{n,m} &= \vec{q}_n \vec{W}_{n-m} \vec{k}_m^T \\
            &= \vec{q}_n^{[0:\frac{d}{2}]} ~\cos((n-m)\vec{\theta})~ (\vec{k}_m^{[0:\frac{d}{2}]})^T \\
                &\quad+ \vec{q}_n^{[\frac{d}{2}:d]} ~\cos((n-m)\vec{\theta})~ (\vec{k}_m^{[\frac{d}{2}:d]})^T \\
                &\quad+ \vec{q}_n^{[0:\frac{d}{2}]} ~\sin((n-m)\vec{\theta})~ (\vec{k}_m^{[\frac{d}{2}:d]})^T\\
                &\quad- \vec{q}_n^{[\frac{d}{2}:d]} ~\sin((n-m)\vec{\theta})~ (\vec{k}_m^{[0:\frac{d}{2}]})^T, \\
            % &= \sum_{i=0}^{d-1} {\cos((n-m)\theta_{i\%\frac{d}{2}})~ q_n^{(i)}~ k_m^{(i)}}
            % + \sum_{i=0}^{d-1} {(-1)^{i\geq \frac{d}{2}}~\sin((n-m)\theta_{i\%\frac{d}{2}})~ q_n^{(i)}~ k_m^{((i+\frac{d}{2})\%\frac{d}{2})}},
            % &= \sum_{i=0, j=i+\frac{d}{2}}^{\frac{d}{2}-1} {\cos((n-m)\theta_i)~ (q_n^{(i)}~ k_m^{(i)} + q_n^{(j)}~ k_m^{(j)})}
            % + \sum_{i=0, j=i+\frac{d}{2}}^{\frac{d}{2}-1} {\sin((n-m)\theta_i)~ (q_n^{(i)}~ k_m^{(j)} - q_n^{(j)}~ k_m^{(i)})},
    \end{aligned}
\end{equation}
where $\vec{q}_n^{[a:b]}$ is the sub-vector of $\vec{q}_n$ from the $a$-th element (inclusive) to the $b$-th element (exclusive). And $\vec{k}_m^{[a:b]}$ are defined similarly.
By defining the trigonometric basis functions:
% (where $\mathrm{diag}(\Vec{x})$ is the function that returns the diagonal elements of the matrix $\Vec{x}$):
% \begin{equation}
%     \vec{\phi}_{n-m} = \begin{pmatrix}
%         \mathrm{diag}(\cos((n-m)\Vec{\theta}))\\
%         \mathrm{diag}(\cos((n-m)\Vec{\theta}))\\
%     \end{pmatrix},
%     \quad\text{and}\quad
%     \vec{\psi}_{n-m} = \begin{pmatrix}
%         \mathrm{diag}(\sin((n-m)\Vec{\theta}))\\
%         \mathrm{diag}(\sin((n-m)\Vec{\theta}))\\
%     \end{pmatrix},
% \end{equation}
% \begin{equation}
%     \phi_{n-m}^{(i)} = \cos((n-m)\theta_{i\%\frac{d}{2}}), \quad\text{and}\quad
%     \psi_{n-m}^{(i)} = (-1)^{i\geq \frac{d}{2}}~\sin((n-m)\theta_{i\%\frac{d}{2}}),
% \end{equation}

\begin{equation}
    \phi_{n-m}^{(i)} = \cos((n-m)\theta_{i\%\frac{d}{2}})\quad
\end{equation}
and:
\begin{equation}
    \quad
    \psi_{n-m}^{(i)} = (-1)^{i\geq \frac{d}{2}}~\sin((n-m)\theta_{i\%\frac{d}{2}})
\end{equation}

% the dot product score $z_{n,m}$ in Eq.~\ref{eq:z_simplified} can be further simplified as follows:
% \begin{equation}
%     \label{eq:z_simplified2}
%     z_{n,m} = \vec{\phi}_{n-m}^T (\Vec{q}_n \odot \vec{k}_m)
%             + \vec{\psi}_{n-m}^T (\Vec{q}_n \odot \tilde{\vec{k}}_m),
% \end{equation}
% where $\odot$ is the element-wise product, and $\tilde{\vec{k}}_m$ is defined as follows:
% \begin{equation}
%     \tilde{\vec{k}}_m = \begin{pmatrix}
%         \vec{k}_m^{[\frac{d}{2}:d]}\\
%         -\vec{k}_m^{[0:\frac{d}{2}]}
%     \end{pmatrix}.
% \end{equation}
% \begin{equation}
%     \phi_{n-m}^{(i)} = \cos((n-m)\theta_{i\%\frac{d}{2}}), \quad\text{and}\quad
%     \psi_{n-m}^{(i)} = (-1)^{i\geq \frac{d}{2}}~\sin((n-m)\theta_{i\%\frac{d}{2}}),
% \end{equation}
Eq.~\ref{eq:z_simplified} can be further simplified as follows:
\begin{equation}
    \label{eq:z_simplified2}
    z_{n,m} = \sum_{i=0}^{d-1} {\phi_{n-m}^{(i)}~ q_n^{(i)}~ k_m^{(i)}}
            + \sum_{i=0}^{d-1} {\psi_{n-m}^{(i)}~ q_n^{(i)}~ k_m^{(i+\frac{d}{2}\%\frac{d}{2})}}.
\end{equation}

Let's model the key vectors $\vec{k}_m$ as a random variable as follows:
\begin{equation}\label{eq:prob_key}
    k_m^{(i)} = \mu_k^{(i)} + \chi_m^{(i)},
\end{equation}
where $\mu_k^{(i)} = E_{m\in[0,N)}[k_m^{(i)}]$ is the mean value of the $i$-th channel of the key vectors over all positions and $\chi_m^{(i)}$ is the random variable with zero mean and variance $\sigma_i^2$.

By substituting the key vectors with the random variable model, the dot product score $z_{n,m}$ in Eq.~\ref{eq:z_simplified2} can be further simplified to two parts, the mean part $\bar{z}_{n,m}$ and the fluctuation part $\tilde{z}_{n,m}$:
\begin{equation}
    z_{n,m} = \bar{z}_{n,m} + \tilde{z}_{n,m},
\end{equation}
where the mean part $\bar{z}_{n,m}$ is
\begin{equation}
    \bar{z}_{n,m} = \sum_{i=0}^{d-1} {\phi_{n-m}^{(i)}~ q_n^{(i)}~ \mu_k^{(i)}}
        + \sum_{i=0}^{d-1} {\psi_{n-m}^{(i)}~ q_n^{(i)}~ \mu_k^{(i+\frac{d}{2}\%\frac{d}{2})}},
\end{equation}
and the fluctuation part $\tilde{z}_{n,m}$ is
\begin{equation}
    \tilde{z}_{n,m} = \sum_{i=0}^{d-1} {\phi_{n-m}^{(i)}~ q_n^{(i)}~ \chi_m^{(i)}}
        + \sum_{i=0}^{d-1} {\psi_{n-m}^{(i)}~ q_n^{(i)}~ \chi_m^{(i+\frac{d}{2}\%\frac{d}{2})}}.
    \label{eq:fluctuation}
\end{equation}

% \subsection{Attention Score after Softmax}
The attention score $a_{n,m}$ is calculated by applying the softmax function to the dot product score $z_{n,m}$ row-wisely:
\begin{equation}
    \label{eq:a}
    a_{n,m} = \frac{\exp(z_{n,m})}{\sum_{j=0}^{L-1} \exp(z_{n,j})},
\end{equation}
where $L$ is the length of the sequence.

% \subsection{Expectation of Dot Product Score}

% \input{tabs/latency_breakdown}

\noindent\textbf{Distribution of queries and keys.}
We assume that the queries and keys are drawn from a random distribution with mean values $E[q_n^{(i)}]$ and $E[k_m^{(i)}]$ and covariances $\sigma_{i,j}$ as follows:
\begin{equation}
    \sigma_{i,j} = E[(q_n^{(i)}-E[q_n^{(i)}])(k_m^{(j)}-E[k_m^{(j)}])].
\end{equation}
The expectation of the product $q_n^{(i)} k_m^{(j)}$ is as follows:
\begin{equation}
    E[q_n^{(i)} k_m^{(j)}] = \mu^2_{i,j} + \sigma_{i,j}.
\end{equation}
where $\mu_{i,j} = E[q_n^{(i)}]E[k_m^{(j)}]$ is the product of the means of $q_n^{(i)}$ and $k_m^{(j)}$.
Thus, the expectation of the dot product $z_{n,m}$ in Eq.~\ref{eq:z_simplified2} is as follows:

% \resizebox{1.06\columnwidth}{!}{
\begin{equation}
\label{eq:exp-nm}
    \begin{aligned}
        &E[z_{n,m}] \\&= 
\sum_{i=0}^{d-1} {\phi_{n-m}^{(i)}~ E[q_n^{(i)} k_m^{(i)}]}
            + \sum_{i=0}^{d-1} {\psi_{n-m}^{(i)}~ E[q_n^{(i)} k_m^{((i+\frac{d}{2})\%\frac{d}{2})}}] \\
&= \sum_{i=0}^{d-1} {\phi_{n-m}^{(i)}~ (\mu^2_{i,i} + \sigma_{i,i})} \\
&\quad+ \sum_{i=0}^{d-1} {\psi_{n-m}^{(i)}~ (\mu^2_{i,(i+\frac{d}{2})\%\frac{d}{2}} + \sigma_{i,(i+\frac{d}{2})\%\frac{d}{2}})}.
    \end{aligned}
\end{equation}
% }

As indicated by Equation \ref{eq:exp-nm}, the expectation of dot product $z_{n, m}$ is a superposition of multiple sinusoidal functions of $(n-m)$.

% \newpage

% \paragraph{Kernel Implementation Details}

% \newpage

\section{Additional Experimental Results}

\subsection{Needle In A Haystack}

\begin{figure}[htp]
    \centering
    \subfloat[NIAH Results of Baseline w/ MInference]{
      \label{sfig:niah_baseline_minference}
      \includegraphics[width=\columnwidth]{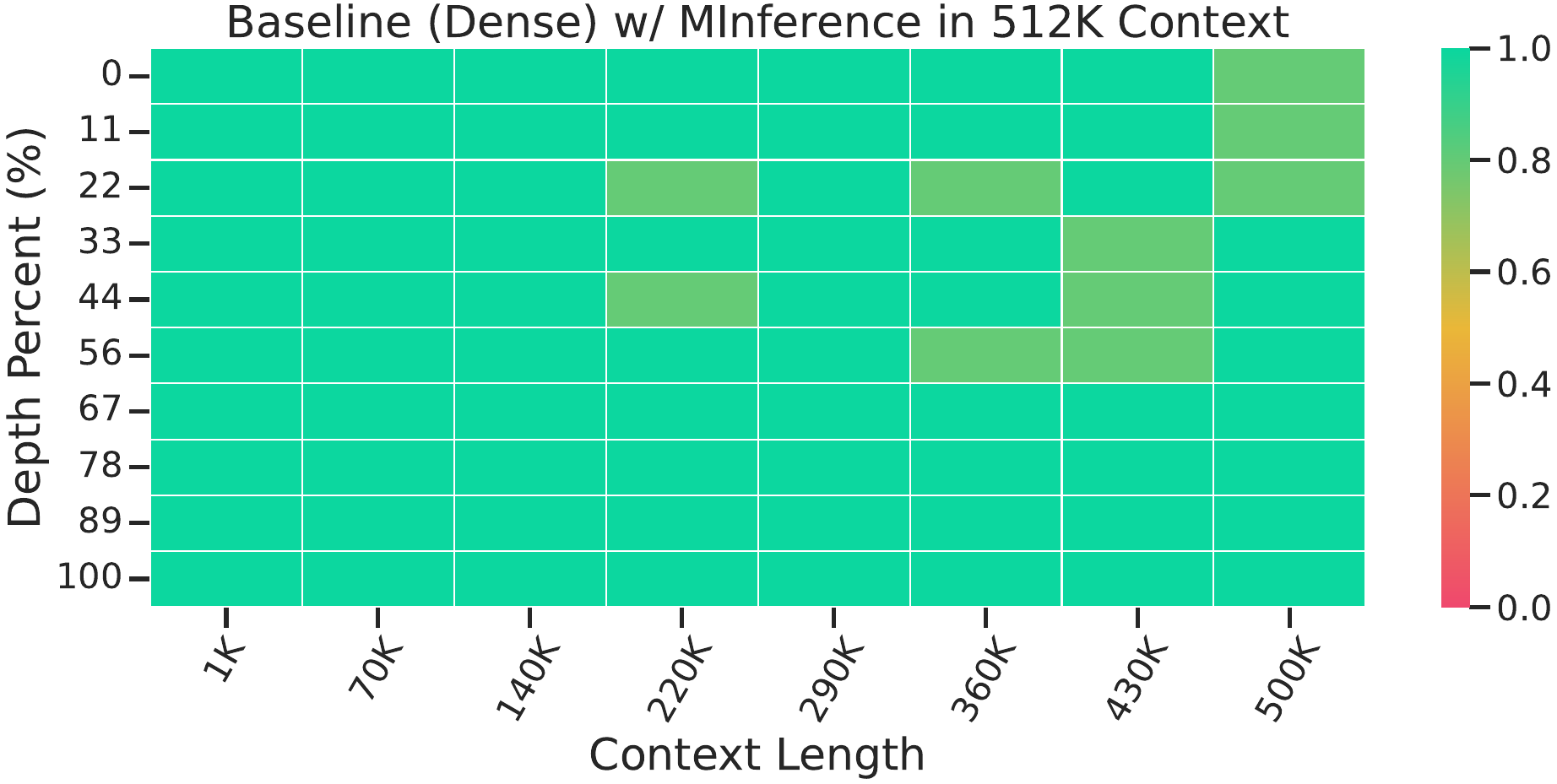}}
      \\
    \subfloat[NIAH Results of MTraining w/ MInference]{
      \label{sfig:niah_mtraining_minference}
      \includegraphics[width=\columnwidth]{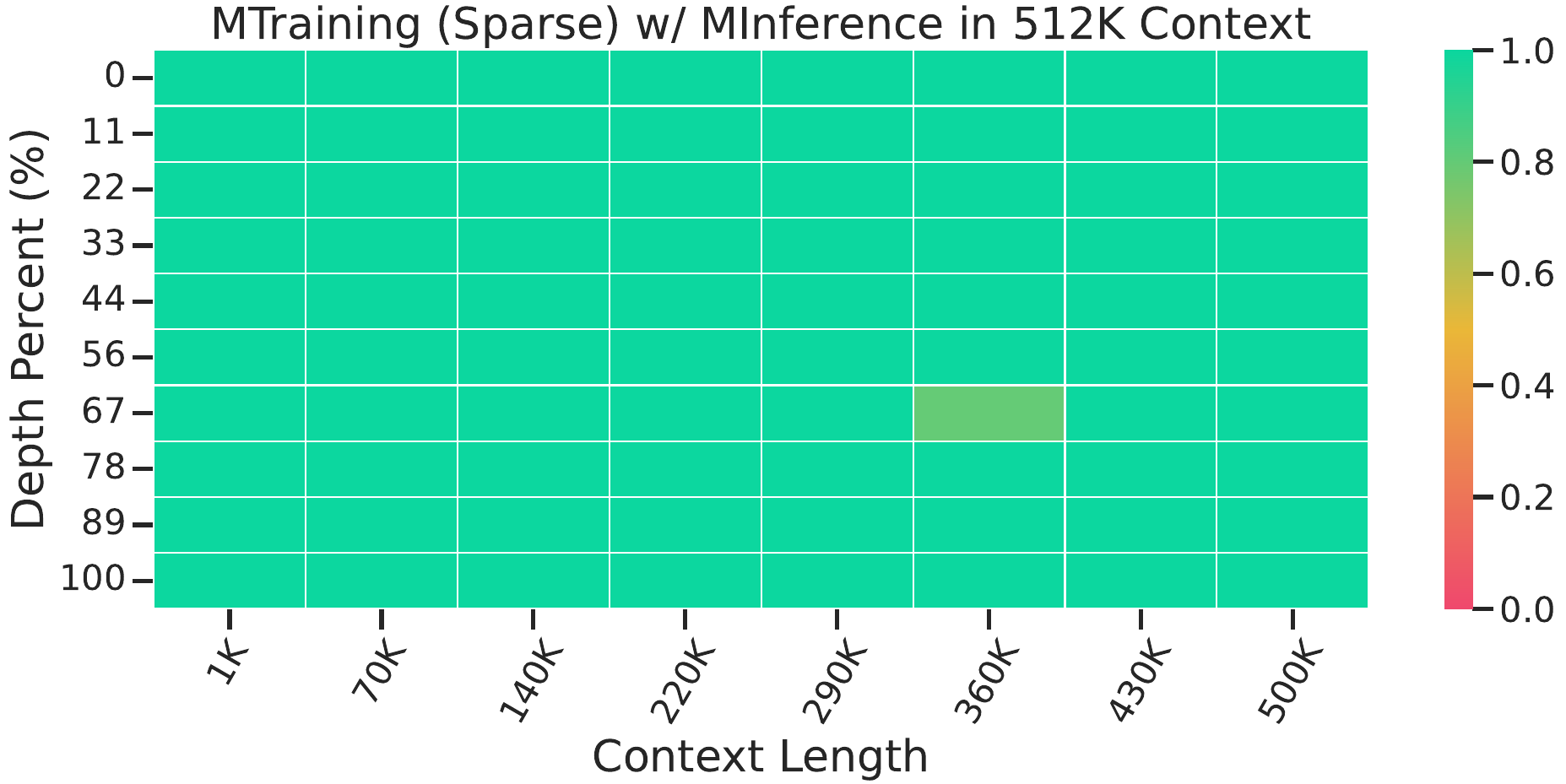}}
    \caption{Needle In A Haystack Results of the baseline checkpoint and the MTraining checkpoint with MInference in the inference stage.}
    \label{fig:niah_results_minference}
    % \vspace{10pt}
  \end{figure}
  
\subsection{Measurement of Workload Imbalance}
\label{app:measure_imbalance}
% To provide a straightforward illustration on how balanced the workload is under different Ring Attention strategies with dynamic sparse attention, we randomly selects a training iteration: we present the distribution of computation time within one Ring Attention step across all 32 devices in Figure \ref{sfig:app_worker_imbalance} and the distribution of computation time for one device across 32 Ring Attention steps in \ref{sfig:app_step_imbalance}. In addition, we also measured the worker- and step-level computation imbalance degree, and the average ratio of computation to training iteration time across all training iterations with different training strategies in Table \ref{tab:imbalance_metrics}.

% \begin{figure*}[tb]
%   % \vspace{-10pt}
%   \centering
%   \subfloat[Worker-level Workload.]{
%     \label{sfig:app_worker_imbalance}
%     \includegraphics[width=\linewidth]{figs/app_worker_imbalance.pdf}}\\
%   \subfloat[Step-level Workload.]{
%     \label{sfig:app_step_imbalance}
%     \includegraphics[width=\linewidth]{figs/app_step_imbalance.pdf}}
%     % \hspace{0.2em}
%     % \hspace{4em}
%   \caption{Distribution of attention computation time using different methods with 512K tokens on 32 GPUs: across CP workers within a fixed Ring Attention step (a) and across Ring Attention steps for a fixed worker (b).}
%   \label{fig:imbalanced_additional}
%   % \vspace{-10pt}
% \end{figure*}

\begin{table*}[h]
  \caption{Average imbalance degree (ID) and Computation Ratio for different training strategies.}
  \centering
  \footnotesize                               % ← switch to a smaller font
  \setlength{\tabcolsep}{4pt}                 % ← tighten column spacing
  \renewcommand{\arraystretch}{1.15}          % ← keep rows comfortable
  \begin{tabular}{@{}p{5cm}ccc@{}}          % ← allow method names to wrap
    \toprule
      & \shortstack[c]{\textbf{Avg.\ ID}\\(Worker-level)} %
      & \shortstack[c]{\textbf{Avg.\ ID}\\(Step-level)}   %
      & \shortstack[c]{\textbf{Avg.\ Comp.}\\Ratio (Step-level)} \\
    \midrule
    \textbf{Dense}                       & $1.02 \pm 0.00$ & $1.01 \pm 0.00$ & $0.88 \pm 0.05$ \\
    \textbf{MTraining }                  & $1.03 \pm 0.02$ & $1.16 \pm 0.01$ & $0.82 \pm 0.00$ \\
    \textbf{MTraining w/o Hierarchical}  & $1.04 \pm 0.01$ & $1.19 \pm 0.04$ & $0.76 \pm 0.01$ \\
    \textbf{MTraining w/ Zigzag}         & $2.61 \pm 0.48$ & $1.99 \pm 0.35$ & $0.42 \pm 0.09$ \\
    \bottomrule
  \end{tabular}
  \label{tab:imbalance_metrics}
\end{table*}

% Table \ref{tab:imbalance_metrics} provides additional numeric measurement of the Imbalance Degree (ID) in the workload distribution over CP workers and across Ring Attention steps.

Table \ref{tab:imbalance_metrics} aggregates statistics across all training iterations, including (i) the average worker-level imbalance degree over all steps and workers, (ii) the average step-level imbalance degree over all workers, and (iii) the average computation time ratio relative to the Ring Attention step latency.

% \newpage
\section{Additional Experimental Details}
\label{app:additional_experimental_details}

\begin{figure}[h]
    \centering
    \includegraphics[width=\linewidth]{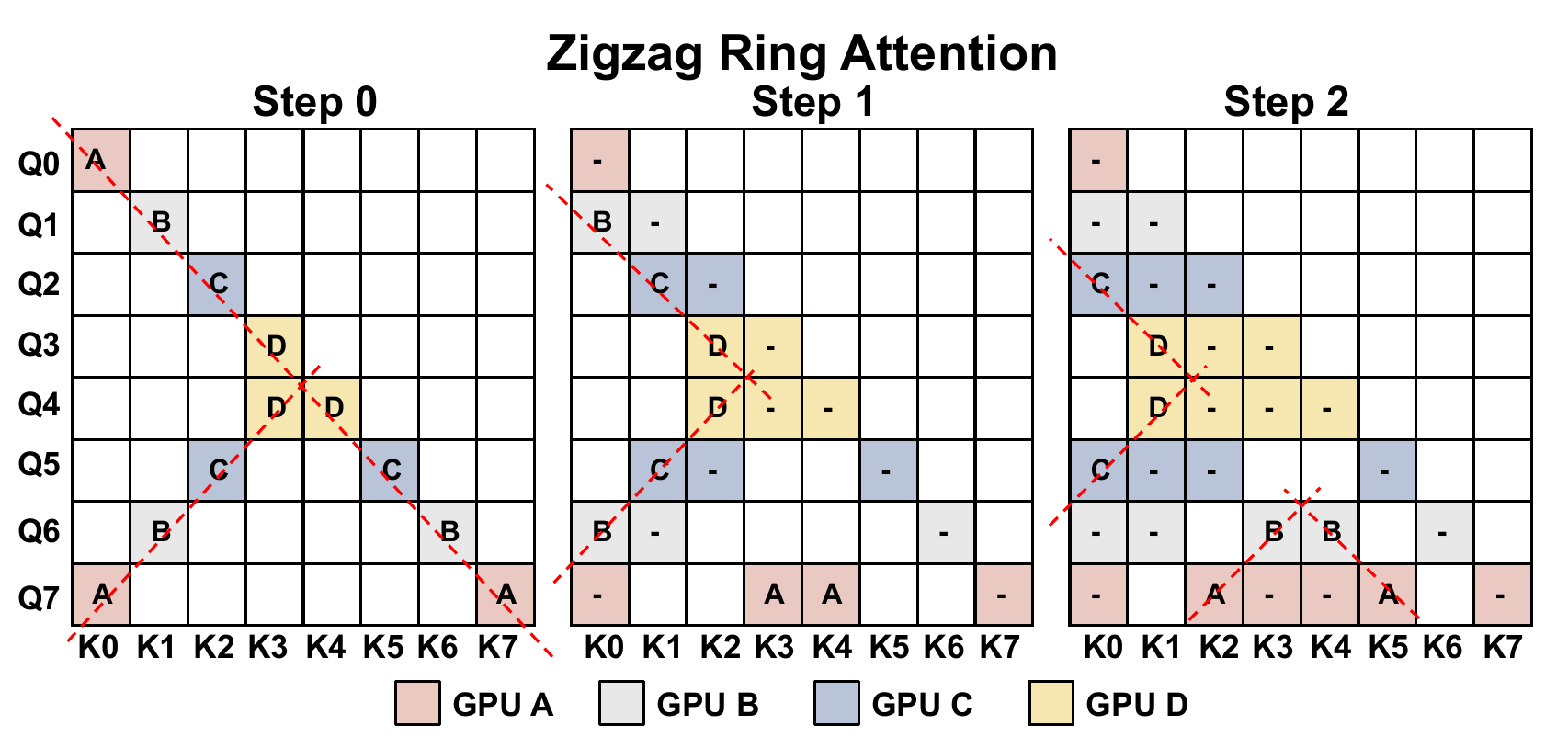}
    \caption{Step-level Computation Schedule of Zigzag Ring Attention.}
    % \vspace{-10pt}
    \label{fig:zigzag_comp}
\end{figure}

\paragraph{ZigZag} Figure \ref{fig:zigzag_comp} provides a visualization of step-level computation schedule of ZigZag Ring Attention, complementing those of Striped Ring Attention and Hierarchical Balanced Sparse Ring Attention in Figure \ref{fig:comp_schedule}. 

\paragraph{Hierarchical Balanced Sparse Ring Attention} The pseudocode of implementation can be found in Algorithm \ref{alg:sparse_double_ring}.

\paragraph{Additional Implementation Details}

All experiments were conducted on a 4 × 8 NVIDIA A100-40 GB cluster, where the eight GPUs inside each node communicate via NVLink and nodes are interconnected through HDR InfiniBand. Because this study isolates the benefits of Context Parallelism, every GPU in both training and profiling runs serves exclusively as a CP worker, with no additional data, pipeline, or tensor parallelism enabled.  We employ the nnScaler framework~\cite{lin2024nnscaler}, which first traces the model into a computation graph and then searches for an optimal parallel execution plan; its search space is constrained so that the resulting plan assigns all GPUs to CP only.  Training uses ZeRO-2~\cite{rajbhandari2020zero}, 64 gradient-accumulation steps~\cite{huang2019gpipe}, bfloat16 precision for model weights, gradients, and activations, and float32 precision for optimiser states; the optimiser is Adam~\cite{kingma2014adam}; gradient checkpointing and recompute~\cite{chen2016training} are applied to peak activation memory. Efficiency-profiling sessions replicate the same parallel-execution configuration. Self-attention in MTraining are implemented with custom CUDA kernels built upon FlashAttention~\cite{dao2022flashattention}, BlockSparse~\cite{guo2024blocksparse}, and the PIT dynamic-sparse compiler~\cite{zheng2023pit}. For external sparse algorithms such as MoBA and XAttention, we adapt their original code to operate under Zigzag Ring-Attention schedule.

\paragraph{Baselines Details}
1) MoBA \cite{lu2025moba}. MoBA partitions the key-value sequence into fixed-size blocks and, for every query, an MoE-style gate chooses the top-k most relevant blocks (always including the query’s own block) before running FlashAttention inside each selected block. In our experiments, the block size is set to 4096 and topK value is 12, making the sparse ratio under 512K context be 0.9. The implementation published in their official repo\footnote{https://github.com/MoonshotAI/MoBA} is adapted to enable it to run with Zigzag Ring Attention. But the efficiency of the officially released code is suboptimal, we ignore the comparison with it in efficiency-related experiments. 

2) XAttention \cite{xu2025xattention}. XAttention score square blocks by summing every certain stride along their antidiagonals and retains only the high-score blocks, giving a plug-and-play, training-free block-sparse attention that accelerates prefill while matching dense accuracy. In our experiments, we use the following settings with granularity being 128 as the block size,  stride 16 as the sampling pitch and threshold: 0.9 for selecting blocks.

\restylefloat{algorithm}

\begin{figure}[htbp]
% \begin{wrapfigure}{r}{0.67\columnwidth}
\centering
% \vspace{-60pt}
% \begin{minipage}[t]{0.9\textwidth}
\vspace{0pt}
\centering
\begin{algorithm}[H]
%   \caption{\method{} is, for the most part, a stateless replacement of the attention operation in a pre-trained transformer. The exception is $\overline{\boldsymbol{v}} = \mathrm{mean}(\boldsymbol{V})$, which should be cached and updated during inference using a running-mean algorithm in order to avoid downloading the entire $\boldsymbol{V}$ matrix at every step. In our experiments, we set $r \in \left\{16, 32, 64\right\}$ ($r$ is the main determinant of the compression ratio), $k \in \left\{64, 128\right\}$ and $l = k/4$. PyTorch code is provided in \cref{sec:app:code}.}
\captionsetup[algorithm]{singlelinecheck=off}
\caption{Balanced Sparse Ring Attention fuse w/ Hierarchical Sparse Ring Attention }
\label{alg:sparse_double_ring}
\begin{algorithmic}
  \STATE {\bfseries World size and rank:} $w_{outer}, w_{inner}, r$
  \STATE {\bfseries Input data:} $Q, K, V$
  \STATE {\bfseries Vertical and slash Index:} $I_v, I_s$

    \LineComment{{\# Convert sparse index for current rank}}
    \STATE $I_{block}, I_{bar} = \mathrm{convert\_index}(I_v, I_s, w_{outer} * w_{inner}, r)$

    \LineComment{{\# Outer ring}}
    
  \FOR{$i \gets 1$ to $w_{outer}$}
    \IF{$i < w_{outer}$}

        \STATE \LineComment{{\# Start outer communication}}

        \STATE $\mathrm{next\_outer\_rank} = (r + w_{inner}) \% (w_{outer} * w_{inner})$
        \STATE $\mathrm{P2P_{outer}.async\_send}(K, \mathrm{next\_outer\_rank})$
        \STATE $\mathrm{P2P_{outer}.async\_send}(V, \mathrm{next\_outer\_rank})$
        \STATE $\mathrm{prev\_outer\_rank} = (r - w_{inner}) \% (w_{outer} * w_{inner})$
        \STATE $K'' = \mathrm{P2P_{outer}.async\_recv}(\mathrm{prev\_outer\_rank})$
        \STATE $V'' = \mathrm{P2P_{outer}.async\_recv}(\mathrm{prev\_outer\_rank})$
    \ENDIF

    \LineComment{{\# Inner ring}}
      \FOR{$j \gets 1$ to $w_{inner}$}
    
        \IF{$j < w_{inner}$}
    
        \STATE \LineComment{{\# Start inner communication}}

            \STATE $\mathrm{next\_inner\_rank} = (r + 1) \% w_{inner}$
            \STATE $\mathrm{P2P_{inner}.async\_send}(K, \mathrm{next\_inner\_rank})$
            \STATE $\mathrm{P2P_{inner}.async\_send}(V, \mathrm{next\_inner\_rank})$
            \STATE $\mathrm{prev\_inner\_rank} = (r - 1) \% w_{inner}$
            \STATE $K' = \mathrm{P2P.async\_recv}(\mathrm{prev\_inner\_rank})$
            \STATE $V' = \mathrm{P2P.async\_recv}(\mathrm{prev\_inner\_rank})$
        \ENDIF

        \LineComment{{\# Sparse attention computation}}
        \STATE $Out', LSE' \gets \mathrm{block\_bar\_sparse\_attention\_forward}($
        \STATE $\quad 
        Q, K, V, I_{block}[i * w_{inner} + j], I_{bar}[i * w_{inner} + j]$
        \STATE$)$
        \STATE $Out, LSE \gets \mathrm{merge\_out\_and\_lse}(Out, LSE, Out', LSE')$

        \IF{$j < w_{inner}$}
    
        \STATE \LineComment{{\# Wait inner communication}}

            \STATE $\mathrm{P2P_{inner}.wait}()$
            \STATE $K \gets K'$, $V \gets V'$
        \ENDIF
      \ENDFOR
    
    \IF{$i < w_{outer}$}

        \STATE \LineComment{{\# Wait outer communication}}

        \STATE $\mathrm{P2P_{outer}.wait}()$
        \STATE $K \gets K''$, $V \gets V''$
    \ENDIF
  \ENDFOR

\end{algorithmic}
\end{algorithm}
% \end{minipage}
% \vspace{-38pt}
% \end{wrapfigure}
\end{figure}

% \section{Poster}

% We propose MTraining, a distributed methodology leveraging dynamic sparse attention to enable efficient training for LLMs with ultra-long contexts. It integrates three key components: a distributed sparse index approximating algorithm, balanced sparse ring attention, and hierarchical sparse ring attention. They are designed to synergistically address:

% - 

% \section{Please add supplemental material as appendix here}
% %
% Put anything that you might normally include after the references as an appendix here, {\it not in a separate supplementary file}. Upload your final camera-ready as a single pdf, including all appendices.

%%%%%%%%%%%%%%%%%%%%%%%%%%%%%%%%%%%%%%%%%%%%%%%%%%%%%%%%%%%%%%%%%%%%%%%%%%%%%%%
%%%%%%%%%%%%%%%%%%%%%%%%%%%%%%%%%%%%%%%%%%%%%%%%%%%%%%%%%%%%%%%%%%%%%%%%%%%%%%%

\end{document}